\documentclass[conference, onecolumn, 10pt, peerreview]{IEEEtran}
\usepackage[dvips]{graphicx}
\usepackage[english]{babel}
\usepackage[latin1]{inputenc}

\usepackage{amsmath}
\usepackage{amssymb}
\usepackage{amsbsy}
\usepackage{url}
\usepackage{pstricks}

\usepackage{cite} 
                        
\newcommand{\betP}{\mathrm{BetP}}
\newcommand{\ind}{\mbox{1\hspace{-.25em}l}}
\newcommand{\mix}{\mathrm{Mix}}

\newcommand{\Y}{\mathrm{Y}}

\newcommand{\Flo}{\mathrm{Flo}}
\newcommand{\PCR}{\mathrm{PCR}}
\newcommand{\DPCR}{\mathrm{DPCR}}
\newcommand{\MDPCR}{\mathrm{MDPCR}}
\newcommand{\PCRmo}{\mathrm{PCR6}}
\newcommand{\PCRdsm}{\mathrm{PCR5}}

\newcommand{\eN}{\mbox{I\hspace{-.15em}N}}

\begin{document}

\title{General combination rules for \\
qualitative and quantitative beliefs\footnote{Manuscript received September 24, 2007; released for publication August 5, 2008. \\
Refereeing of this contribution was handled by Fabio Roli.}} 
\author{\authorblockN{Arnaud Martin, Christophe Osswald}
\authorblockA{$\mbox{E}^3\mbox{I}^2$ EA3876\\
ENSIETA\\
2 rue Fran{\c c}ois Verny,\\
29806 Brest Cedex 09, France.\\
Email:Arnaud.Martin,Christophe.Osswald@ensieta.fr} \and
\authorblockN{\\Jean Dezert}
\authorblockA{
ONERA\\
The French Aerospace Lab\\
29 Av. Division Leclerc,\\
92320 Ch\^atillon, France.\\
Email:jean.dezert@onera.fr}\and
\authorblockN{\\Florentin Smarandache}
\authorblockA{\\Chair of Math. \& Sciences Dept.\\
University of New Mexico,\\
200 College Road,\\
Gallup, NM 87301, U.S.A.\\
Email: smarand@unm.edu}}

\maketitle

\selectlanguage{english}

\begin{abstract}
Martin and Osswald \cite{Martin07} have recently proposed many generalizations of combination rules on quantitative beliefs in order to manage the conflict and to consider the specificity of the responses of the experts. Since the experts express themselves usually in natural language with linguistic labels, Smarandache and Dezert \cite{Li07} have introduced a mathematical framework for dealing directly also with qualitative beliefs.  In this paper we recall some element of our previous works and propose the new combination rules, developed for the fusion of both qualitative or quantitative beliefs.\\
%
\end{abstract}


%
\IEEEpeerreviewmaketitle

\section{Introduction}
\label{sec:Introduction}
Many fusion theories have been studied for the combination of the experts opinions expressed either quantitatively or qualitatively such as voting rules \cite{Xu92,Lam97}, possibility theory \cite{Zadeh78,Dubois88a}, and belief functions theory \cite{Dempster67,Shafer76}. All these fusion approaches can be divided basically into four steps: \textit{modelization}, \textit{parameters estimation} (depending on the model, not always necessary)), \textit{combination} and \textit{decision}. The most difficult step is presumably the first one which depends highly on the problem and application we have to cope with. However, it is only at the combination step that we can take into account useful information such as the conflict (partial or total) between the experts and/or the specificity of the expert's response. 

The voting rules are not adapted to the modelization of conflict between experts \cite{Xu92}. Although both possibility and probability-based theories can model imprecise and uncertain data at the same time, in many applications, the experts are only able to express their ``certainty'' (or belief) only from their partial knowledge, experience and from their own perception of the reality. In such context, the belief function-based theories provide an appealing general mathematical framework for dealing with quantitative and qualitative beliefs.

In this paper we present the most recent advances in belief functions theory for managing the conflict between the sources of evidence/experts and their specificity. For the first time in the literature both the quantitative and qualitative aspects of beliefs are presented in a unified mathematical framework. This paper actually extends two papers \cite{Li07,Martin07} presented during the 10th International Conference on Information Fusion (Fusion 2007) in Qu\' ebec City, Canada on July 9-12, 2007 in the session ``Combination in Evidence Theory''.\\

Section \ref{Sec2} briefly recalls the basis of belief functions theories, \emph{i.e.} the Mathematical Theory of Evidence or Dempster-Shafer theory (DST) developed by Shafer in 1976 \cite{Dempster67, Shafer76}, and its natural extension called Dezert-Smarandache Theory (DSmT) \cite{DSmTBook1,DSmTBook2}.
We introduce in this section the notion of quantitative and qualitative beliefs and the operators on linguistic labels for dealing directly with qualitative beliefs. Section \ref{Sec3} presents the main classical quantitative combination rules used so far, \emph{i.e.} Dempster's rule, Yager's rule, Dubois-Prade's rule and the recent Proportional Conflict Redistribution rules (PCR) proposed by Smarandache and Dezert \cite{Smarandache06} and extended by Martin and Osswald in \cite{DSmTBook2}. Some examples are given to illustrate how these rules work. Section \ref{Sec4} explains through different examples how all the classical quantitative combination rules can be directly and simply translated/extended into the qualitative domain in order to combine easily any qualitative beliefs expressed in natural language by linguistic labels. Section \ref{Sec5} proposes new general quantitative rules of combination which allow to take into account both the discounting of the sources (if any) and the proportional conflict redistribution. The direct extension of these general rules into the qualitative domain is then presented in details on several examples in section \ref{Sec6}. 

\section{Basis of DST and DSmT}
\label{Sec2}

\subsection{Power set and hyper-power set}

In DST framework, one considers a frame of discernment $\Theta=\{\theta_{1},\ldots,\theta_{n}\}$ as a finite set of $n$ exclusive and exhaustive elements (\emph{i.e.} Shafer's model denoted $\mathcal{M}^0(\Theta)$). The {\it{power set}} of $\Theta$ is the set of all subsets of $\Theta$. The order of a power set of a set of order/cardinality $\vert\Theta\vert=n$ is $2^n$. The power set of  $\Theta$ is denoted $2^\Theta$.  For example, if $\Theta=\{\theta_1,\theta_2\}$, then $2^\Theta=\{\emptyset,\theta_1,\theta_2,\theta_1\cup\theta_2\}$.

In DSmT framework, one considers $\Theta=\{\theta_{1},\ldots,\theta_{n}\}$ be a finite set of $n$ exhaustive elements only (\emph{i.e.} free DSm-model denoted $\mathcal{M}^f(\Theta))$. Eventually some integrity constraints can be introduced in this free model depending on the nature of the problem of interest. The {\it{hyper-power set}} of $\Theta$ (\emph{i.e.} the free Dedekind's lattice) denoted $D^\Theta$ \cite{DSmTBook1} is defined as: 
\begin{enumerate}
\item $\emptyset, \theta_1,\ldots, \theta_n \in D^\Theta$.
\item  If $A, B \in D^\Theta$, then $A\cap B$, $A\cup B \in D^\Theta$.
\item No other elements belong to $D^\Theta$, except those obtained by using rules 1 or 2.
\end{enumerate}
If $\vert\Theta\vert=n$, then $\vert D^\Theta\vert \leq 2^{2^n}$. Since for any finite set $\Theta$, $\vert D^\Theta\vert  \geq \vert 2^\Theta\vert $, we call $D^\Theta$ the  {\it{hyper-power set}} of $\Theta$.  For example, if $\Theta=\{\theta_1,\theta_2\}$, then $D^\Theta=\{\emptyset,\theta_1\cap\theta_2,\theta_1,\theta_2,\theta_1\cup\theta_2\}$. The {\it{free DSm model}} $\mathcal{M}^f(\Theta)$ corresponding to $D^\Theta$ allows to work with vague concepts which exhibit a continuous and relative intrinsic nature. Such concepts cannot be precisely refined in an absolute interpretation because of the unreachable universal truth.

It is clear that Shafer's model $\mathcal{M}^0(\Theta)$ which assumes that all elements of $\Theta$ are truly exclusive is a more constrained model than the free-DSm model $\mathcal{M}^f(\Theta)$ and the power set $2^\Theta$ can be obtained from hyper-power set $D^\Theta$ by introducing in $\mathcal{M}^f(\Theta)$ all exclusivity constraints between elements of $\Theta$. Between the free-DSm model $\mathcal{M}^f(\Theta)$ and Shafer's model $\mathcal{M}^0(\Theta)$, there exists a wide class of fusion problems represented in term of the DSm hybrid models denoted $\mathcal{M}(\Theta)$ where $\Theta$ involves both fuzzy continuous and discrete hypotheses. The main differences between DST and DSmT frameworks are (i) the model on which one works with, and (ii) the choice of the combination rule and conditioning rules \cite{DSmTBook1,DSmTBook2}. In the sequel, we use the generic notation $G^\Theta$ for denoting either $D^\Theta$ (when working in DSmT with free DSm model) or $2^\Theta$ (when  working in DST with Shafer's model). 

\subsection{Quantitative basic belief assignment (bba)}

The (quantitative) basic belief assignment (bba) $m(.)$ has been introduced for the first time in 1976 by Shafer \cite{Shafer76} in his Mathematical Theory of Evidence (\emph{i.e.} DST). $m(.)$ is defined as a mapping function from $2^\Theta \rightarrow [0,1]$ provided by a given source of evidence $\mathcal{B}$ satisfying the conditions:
\begin{equation}
m(\emptyset)=0, 
\label{close}
\end{equation}

\begin{equation}
\sum_{A\in 2^\Theta} m(A) = 1.
\label{normDST}
\end{equation}

The elements of $2^\Theta$ having a strictly positive mass are called {\it{focal elements}} of $\mathcal{B}$. The set of focal elements of $m(.)$ is called the core of $m(.)$ and is usually denoted $\mathcal{F}(m)$. The equation (\ref{close}) corresponds to the closed-world assumption \cite{Shafer76}. As introduced by Smets \cite{Smets90}, we can also define the belief function only with:
\begin{equation}
\label{open}
\sum_{A\in 2^\Theta} m(A) = 1.
\end{equation}
\noindent and thus we can have $m(\emptyset)>0$, working with the open-world assumption. In order to change an open world to a closed world, we can always add one extra closure element in the open discriminant space $\Theta$. In the following, we assume that we always work within a closed-world $\Theta$.

The (quantitative) basic belief assignment (bba) $m(.)$ can also be defined similarly in the DSmT framework by working on hyper-power set $D^\Theta$ instead on classical power-set $2^\Theta$ as within DST. More generally for taking into account some integrity constraints on (closed-world) $\Theta$ (if any), $m(.)$ can be defined on $G^\Theta$ as:

\begin{equation}
m(\emptyset)=0,
\label{closeDSmT}
\end{equation}

\begin{equation}
\sum_{A\in G^\Theta} m(A) = 1.
\label{normDSmT}
\end{equation}

The conditions \eqref{close}-\eqref{normDSmT} give a large panel of definitions of the belief functions, which is one of the difficulties of the theories. From any basic belief assignments $m(.)$, other belief functions can be defined such as the credibility $\text{Bel}(.)$ and the plausibility $\text{Pl}(.)$ \cite{Shafer76,DSmTBook1} which are in one-to-one correspondence with $m(.)$.\\

After combining several bba's provided by several sources of evidence into a single one with some chosen fusion rule (see next section), one usually has also to make a final decision to select the ``best'' hypothesis representing the unknown truth for the problem under consideration. 
%
%
Several approaches are generally adopted for decision-making from belief functions $m(.)$, $\text{Bel}(.)$ or $\text{Pl}(.)$. The maximum of the credibility function $\text{Bel}(.)$ is known to provide a pessimistic decision, while the maximum of the plausibility function $\text{Pl}(.)$ is often considered as too optimistic. A common solution for decision-making in these frameworks is to use the {\it{pignistic probability}} denoted $\betP(X)$ \cite{Smets90} which offers a good compromise between the max of $\text{Bel}(.)$ and the max of  $\text{Pl}(.)$. The pignistic probability in DST framework is given for all $X \in 2^\Theta$, with $X \neq \emptyset$ by:
\begin{eqnarray}
\label{pignistic}
\betP(X)=\sum_{Y \in 2^\Theta, Y \neq \emptyset} \frac{|X \cap Y|}{|Y|} \frac{m(Y)}{1-m(\emptyset)},
\end{eqnarray}
for $m(\emptyset)\neq 1$.
\noindent
\noindent
The pignistic probability can also be defined in DSmT framework as well (see Chapter 7 of \cite{DSmTBook1} for details).\\

When we can quantify/estimate the reliability of each source of evidence, we can weaken the basic belief assignment before the combination by the classical discounting procedure \cite{Shafer76}:
\begin{eqnarray}
\label{massDisounted}
\left\{
\begin{array}{l}
	m'(X)=\alpha m(X), \, \forall X \in 2^\Theta \setminus \{\Theta\} \\
	m'(\Theta)=\alpha m(\Theta)+1-\alpha.
\end{array}
\right.
\end{eqnarray}
$\alpha \in [0,1]$ is the discounting factor of the source of evidence $\mathcal{B}$ that is in this case the reliability of the source of evidence $\mathcal{B}$, eventually as a function of $X \in 2^\Theta$. Same procedure can be applied for bba's defined on $G^\Theta$ in DSmT framework.

\subsection{Qualitative basic belief assignment}

\subsubsection{Qualitative operators on linguistic labels}
\label{QualitativeOperators}

Recently Smarandache and Dezert \cite{DSmTBook2,DSmTBook3,Li07} have proposed an extension of classical quantitative belief assignments and numerical operators to qualitative beliefs expressed by linguistic labels and qualitative operators in order to be closer to what human experts can easily provide. In order to compute directly with words/linguistic labels and qualitative belief assignments instead of quantitative belief assignments over $G^\Theta$, Smarandache and Dezert have defined in \cite{DSmTBook2} a {\it{qualitative basic belief assignment}} $qm(.)$ as a mapping function from $G^\Theta$ into a set of linguistic labels $L=\{L_0,\tilde{L},L_{n+1}\}$ where $\tilde{L}=\{L_1,\cdots,L_n\}$
is a finite set of linguistic labels and where $n \ge 2$ is an
integer. For example, $L_1$ can take the linguistic value ``poor'',
$L_2$ the linguistic value ``good'', etc. $\tilde{L}$ is endowed with
a total order relationship $\prec$, so that $L_1 \prec L_2 \prec
\cdots \prec L_n $. To work on a true closed linguistic set $L$
under linguistic addition and multiplication operators, Smarandache
and Dezert extended naturally $\tilde{L}$ with two extreme values
$L_0=L_{\min}$ and $ L_{n+1}=L_{\max}$, where $L_0$ corresponds to the minimal
qualitative value and $L_{n+1}$ corresponds to the maximal
qualitative value, in such a way that $L_0 \prec L_1 \prec L_2 \prec
\cdots \prec L_n  \prec L_{n+1} $, where $\prec$ means inferior to, less (in quality) than, or smaller than, etc.  Labels $L_0$, $L_1$, $L_2$, \ldots, $L_n$, $L_{n+1}$ are called {\it{linguistically equidistant}} if:
$L_{i+1} - L_i = L_i - L_{i-1}$ for all $i = 1, 2, \ldots, n$ where the definition of subtraction of labels is given in  the sequel by \eqref{eq:qsub}. In the sequel $L_i \in L$ are assumed linguistically equidistant\footnote{If the labels are not equidistant, the q-operators still work, but they are less accurate.} labels such that we can make an isomorphism between $L = \{L_0, L_1, L_2, \ldots, L_n, L_{n+1}\}$ and $\{ 0, 1/(n+1), 2/(n+1), \ldots, n/(n+1), 1\}$, defined as $L_i = i/(n+1)$ for all $i = 0, 1, 2,\ldots, n, n+1$. Using this isomorphism, and making an analogy to the classical operations of real numbers, we are able to justify and define precisely the following qualitative operators (or $q$-operators for short):

\begin{itemize}
\item $q$-addition of linguistic labels:
\begin{equation}
L_i+L_j=\frac{i}{n+1}+ \frac{j}{n+1}=\frac{i+j}{n+1}=L_{i+j},
\label{eq:q-addition}
\end{equation}
\noindent we set the restriction that $i+j < n+1$; in the case when $i+j \geq n+1$ we restrict $L_{i+j} = L_{n+1}=L_{\max}$. This is the justification of the qualitative addition we have defined. 

\item $q$-multiplication of linguistic labels\footnote{The $q$-multiplication of two linguistic labels defined here can be extended directly to the multiplication of $n>2$ linguistic labels. For example the product of three linguistic label will be defined as $L_i \cdot L_j  \cdot L_k = L_{[(i\cdot j\cdot k)/(n+1)(n+1)]}$, etc.}:

\begin{itemize}
\item[a)] Since $L_i \cdot L_j= \frac{i}{n+1}\cdot  \frac{j}{n+1} = \frac{(i\cdot j)/(n+1)}{n+1}$, the best approximation would be $L_{[(i\cdot j)/(n+1)]}$, where $[x]$ means the closest integer to $x$ (with $[n+0.5]=n+1$,  $\forall n\in \eN$), \emph{i.e.}
\begin{equation}
L_i \cdot L_j = L_{[(i\cdot j)/(n+1)]}.
\label{eq:qmult}
\end{equation}
For example, if we have $L_0$, $L_1$, $L_2$, $L_3$, $L_4$, $L_5$, corresponding to respectively $0$, $0.2$, $0.4$, $0.6$, $0.8$, $1$, then $L_2 \cdot L_3 = L_{[(2\cdot 3)/5]} = L_{[6/5]} = L_{[1.2] }= L_1$; using numbers: $0.4\cdot 0.6 = 0.24 \approx 0.2 = L_1$; also $L_3 \cdot L_3 = L_{[(3\cdot 3)/5]} = L_{[9/5]} = L_{[1.8]} = L_2$; using numbers $0.6 \cdot 0.6 = 0.36 \approx 0.4 = L_2$.
\item[b)] A simpler approximation of the multiplication, but less
accurate (as proposed in \cite{DSmTBook2}) is thus:
\begin{equation}
L_i \cdot L_j = L_{\min\{i,j\}}. \label{eq:qmultsimpler}
\end{equation}
\end{itemize}

\item Scalar multiplication of a linguistic label:

Let $a$ be a real number. We define the multiplication of a linguistic label by a scalar as follows:

\begin{equation}
a \cdot L_i =\frac{a\cdot i}{n+1} \approx
\begin{cases}
L_{[a\cdot i]} & \text{if} \ [a\cdot i]\geq  0,\\
L_{-[a\cdot i]} & \text{otherwise}.
\end{cases}
\label{eq:sqmult}
\end{equation}

\item Division of linguistic labels:

\begin{itemize}
\item[a)]  Division as an internal operator:  $/ : L \cdot  L \rightarrow  L$.
Let $j\neq 0$, then
\begin{equation}
L_i / L_j  =
\begin{cases}
L_{[(i/j)  \cdot (n+1)]} & \text{if} [(i/j) \cdot (n+1)] < n+1,\\
L_{n+1} & \text{otherwise}.
\end{cases}
\label{eq:sqdiv}
\end{equation}

The first equality in \eqref{eq:sqdiv} is well justified because when $ [(i/j) \cdot (n+1)] < n+1$, one has: 
$$L_i / L_j  = \frac{i/(n+1)}{j/(n+1)}=\frac{(i/j)\cdot (n+1)}{n+1}=L_{[(i/j)  \cdot (n+1)]}.$$
For example, if we have $L_0$, $L_1$, $L_2$, $L_3$, $L_4$, $L_5$, corresponding to respectively $0$, $0.2$, $0.4$, $0.6$, $0.8$, $1$, then: $L_1 / L_3 = L_{[(1/3)\cdot  5]}=L_{[5/3]}=L_{[1.66]} \approx  L_2$. $L_4 / L_2 = L_{[(4/2)\cdot 5]}=L_{[2\cdot 5]}=L_{\max}=L_5$ since $10>5$.

\item[b)] Division as an external operator: $\oslash : L \cdot  L \rightarrow  \mathbb{R}^+$.
Let $j\neq 0$. Since $L_i \oslash L_j = (i/(n+1)) / (j/(n+1)) = i/j$, we simply define: 
\begin{equation}
L_i \oslash L_j  = i/j.
\label{eq:sqextdiv}
\end{equation}
\noindent
Justification of b): When we divide say $L_4 / L_1$ in the above example, we get $0.8/0.2 = 4$, but no label is corresponding to number 4 which is not included in the interval $[0,1]$, hence the division as an internal operator we need to get as a response label, so in our example we approximate it to $L_{\max}=L_5$, which is a very rough approximation! Therefore, depending on the fusion combination rules, it may be better to consider the qualitative division as an external operator, which gives us the exact result.
\end{itemize}

\item $q$-subtraction of linguistic labels:  $- : L \cdot  L \rightarrow \{L, -L\}$, 

\begin{equation}
 L_i - L_j =
\begin{cases}
L_{i-j} & \text{if} \quad i \geq  j,\\
- L_{j-i} &  \text{if} \quad i <  j.
\end{cases}
\label{eq:qsub}
\end{equation}
\noindent
where $ -L =  \{ -L_1, -L_2, \ldots, -L_n, -L_{n+1} \}$. 
The $q$-subtraction above is well justified since when $i \geq  j$, one has $L_i - L_j =  \frac{i}{n+1} - \frac{j}{n+1} =  \frac{i-j}{n+1}$.

\end{itemize}

The previous qualitative operators are logical due to the isomorphism between the set of linguistic equidistant labels and a set of equidistant numbers in the interval $[0,1]$.  These qualitative operators are built exactly on the track of their corresponding numerical operators, so they are more mathematically defined than the ad-hoc definitions of qualitative operators proposed in the literature so far. 
The extension of these operators for handling quantitative or qualitative enriched linguistic labels can be found in \cite{Li07}.\\

\noindent
{\it{Remark about doing multi-operations on labels}}: \\

When working with labels, no matter how many
operations we have, the best (most accurate) result is obtained if we do only one approximation. That one should be at the end. For example, if we have to compute terms like $L_iL_jL_k / (L_p+L_q)$ as for Qualitative Proportional Conflict
Redistribution (QPCR) rule (see example in section \ref{classical qualitative combination_rules}), we compute all operations as defined above. Without any approximations (\emph{i.e.} not even calculating the integer part of indexes, neither replacing by $n+1$ if the intermediate results is bigger than $n+1$), so:


\begin{equation}
\frac{L_iL_jL_k }{L_p+L_q}=\frac{L_{(ijk)/(n+1)^2}}{L_{p+q} }= L_{{ \frac{(ijk) / {(n+1)^2}}{p+q}\cdot (n+1)}} = L_{ \frac{(ijk)/(n+1)}{p+q} } =L_{ \frac{ijk}{(n+1)(p+q)}},
\label{eq12}
\end{equation}

\noindent and now, when all work is done, we compute the integer part of the index, \emph{i.e.} $[\frac{ijk}{(n+1)(p+q)}]$ or replace it by $n+1$ if the final result is bigger than $n+1$.
Therefore, the term $L_iL_jL_k / (L_p+L_q)$ will take the linguistic value $L_{n+1}$ whenever $[\frac{ijk}{(n+1)(p+q)}] > n+1$. This method also insures us of a unique result, and it is mathematically closer to the result that would be obtained if working with corresponding numerical masses. Otherwise, if one approximates either at the beginning or end of each operation or in the middle of calculations, the inaccuracy propagates (becomes bigger) and we obtain different results, depending on the places where the approximations were done. If we need to round the labels' indexes to integer indexes, for a better accuracy of the result, this rounding must be done at the very end. If we work with fractional/decimal indexes (therefore no approximations), then we can normally
apply the qualitative operators one by one in the order they are needed; in this way the quasi-normalization is always kept.\\

\subsubsection{Quasi-normalization of $qm(.)$}

There is no known way to define a normalized $qm(.)$, but a qualitative quasi-normalization \cite{DSmTBook2,Smarandache07} is nevertheless possible when considering equidistant linguistic labels because in such case, $qm(X_i) = L_i$, is equivalent to a quantitative mass $m(X_i) = i/(n+1)$ which is normalized if: 
$$\sum_{X\in G^\Theta} m(X)= \sum_{k} i_k/(n+1)=1,$$
\noindent
but this one is equivalent to: 
$$\sum_{X\in G^\Theta} qm(X)= \sum_{k} L_{i_k}=L_{n+1}.$$
\noindent
In this case, we have a {\it{qualitative normalization}}, similar to the (classical) numerical normalization. However, if the previous labels $L_0$, $L_1$, $L_2$, $\ldots$, $L_n$, $L_{n+1}$ from the set $L$ are not equidistant, the interval $[0, 1]$ cannot be split into equal parts according to the distribution of the labels. Then it makes sense to consider a {\it{qualitative quasi-normalization}}, \emph{i.e.} an approximation of the (classical) numerical normalization for the qualitative masses in the same way:
 $$\sum_{X\in G^\Theta} qm(X)=L_{n+1}.$$
\noindent
In general, if we don't know if the labels are equidistant or not, we say that a qualitative mass is quasi-normalized when the above summation holds. In the sequel, for simplicity, one assumes to work with quasi-normalized qualitative basic belief assignments.\\

From these very simple qualitative operators, it is possible to extend directly all the quantitative combination rules to their qualitative counterparts as we will show in the sequel.\\

\subsubsection{Working with refined labels}

\begin{itemize}
\item We can further extend the standard labels (those
with positive integer indexes) to refined labels, \emph{i.e.}
labels with fractional/decimal indexes.  In such a
way, we get a more exact result, and the quasi-normalization is kept.\\

Consider a simple example: If $L_2 = \text{\it{good}}$ and $L_3 = \text{\it{best}}$, then $L_{2.5} = \text{\it{better}}$,
which is a qualitative (a refined label) in between $L_2$ and $L_3$.

\item
Further, we consider the confidence degree in
a label, and give more interpretations/approximations
to the qualitative information.\\

For example: $L_{2/5} = (1/5)\cdot L_2$, which means that we are 20\% confident in label $L_2$;
 or $L_{2/5} = (2/5)\cdot L_1$, which means that we are 40\% confident in label $L_1$, so $L_1$ is closer to reality than $L_2$;
we get $100\%$ confidence in $L_{2/5} = 1\cdot L_{2/5}$.\\

\end{itemize}

\subsubsection{Working with non-equidistant labels}

We are not able to find (for non-equidistant labels) exact corresponding numerical values in the interval $[0, 1]$ in order to reduce the qualitative fusion to a quantitative fusion, but only approximations.  We, herfore, prefer the use of labels. 

\section{Classical quantitative combination rules}
\label{combination_rules}
\label{Sec3}

The normalized conjunctive combination rule also called Dempster-Shafer (DS) rule is the first rule proposed in the belief theory by Shafer following Dempster's works in sixties \cite{Dempster67}. In the belief functions theory one of the major problems is the conflict repartition enlightened by the famous Zadeh's example \cite{Zadeh_1979}. Since Zadeh's paper, many combination rules have been proposed, building a solution to this problem \cite{Yager87, Dubois88, Smets90b, Inagaki91, Smets93, Josang03, Smarandache05, Florea06, Martin06b, Denoeux06}. In recent years, some unification rules have been proposed \cite{Smets97, Lefevre02a, Appriou05}. We briefly browse the major rules developed and used in the fusion community working with belief functions through last past thirty years (see  \cite{Sentz_2002} and \cite{DSmTBook2} for a more comprehensive survey).

To simplify the notations, we consider only two independent sources of evidence $\mathcal{B}_1$ and $\mathcal{B}_2$ over the same frame $\Theta$ with their corresponding bba's $m_1(.)$ and $m_2(.)$. 
Most of the fusion operators proposed in the literature use either the conjunctive operator, the disjunctive operator or a particular combination of them. These operators are respectively defined $\forall A\in G^\Theta$, by:
\begin{equation}
\label{eq:comb_disj} 
m_{\vee}(A)=(m_1 \vee m_2) (A) =  \sum_{\substack{X,Y\in G^\Theta\\ X \cup Y = A}} m_1(X)m_2(Y),
\end{equation}
\begin{equation}
\label{eq:comb_conj}
m_{\wedge}(A)=(m_1 \wedge m_2) (A) =  \sum_{\substack{X,Y\in G^\Theta\\ X \cap Y = A}} m_1(X)m_2(Y).
\end{equation}
\noindent
The global/total {\it{degree of conflict}} between the sources $\mathcal{B}_1$ and $\mathcal{B}_2$ is defined by :
 \begin{equation}
 k\triangleq m_{\wedge}(\emptyset)=\displaystyle
\sum_{\substack{X,Y\in G^\Theta\\ X\cap Y=\emptyset}} m_{1}(X) m_{2}(Y).
 \end{equation}
 \noindent
If $k$ is close to $0$, the bba's $m_1(.)$ and $m_2(.)$ are almost not in conflict, while if $k$ is close to $1$, the bba's are almost in total conflict. Next, we briefly review the main common quantitative fusion rules encountered in the literature and used in engineering applications.\\


\noindent{\it{Example 1}}: Let's consider the 2D frame $\Theta=\{A,B\}$ and two experts providing the following quantitative belief assignments (masses) $m_1(.)$ and $m_2(.)$:

 \begin{table}[h]
\centering
 \begin{tabular}{|l|c|c|c|}
    \hline
    & $A$ & $B$ & $A\cup B$ \\
    \hline
    $m_1(.)$  & $1/6$ & $3/6$ & $2/6$ \\
    \hline
   $m_2(.)$  & $4/6$ & $1/6$ & $1/6$ \\
     \hline
  \end{tabular}
  \caption{Quantitative inputs for example 1}
\label{TableExample1}
\end{table}
The disjunctive operator yields the following result:

\begin{align*}
m_{\vee}(A) & = m_{1}(A)m_2(A) = (1/6)\cdot (4/6) =4/36
\end{align*}
\begin{align*}
m_{\vee}(B) & = m_{1}(B)m_2(B) = (3/6)\cdot (1/6) = 3/36
\end{align*}
%

\begin{align*}
m_{\vee}(A\cup B) & =  m_{1}(A)m_2(B) +  m_{1}(B)m_2(A) \\
&  \qquad + m_{1}(A)m_{2}(A\cup B)  + m_{2}(A)m_{1}(A\cup B)\\
&  \qquad +   m_{1}(B)m_{2}(A\cup B) + m_{2}(B)m_{1}(A\cup B) \\
&  \qquad + m_{1}(A\cup B)m_{2}(A\cup B)\\
& = (1/6)\cdot (1/6) + (3/6)\cdot (4/6)\\
& \qquad + (1/6)\cdot (1/6) + (4/6)\cdot (2/6)\\
& \qquad +  (3/6)\cdot (1/6) + (1/6)\cdot (2/6)\\
& \qquad + (2/6)\cdot (1/6)\\
&= 29/36
\end{align*}

\noindent
while the conjunctive operator yields:
%
\begin{align*}
m_{\wedge}(A) & = m_{1}(A)m_2(A) + m_{1}(A)m_2(A\cup B)  + m_{2}(A)m_1(A\cup B) \\
&= (1/6)\cdot (4/6) + (1/6)\cdot (1/6) +  (4/6)\cdot (2/6) = 13/36
\end{align*}
\begin{align*}
 m_{\wedge}(B)& = m_{1}(B)m_2(B) + m_{1}(B)m_2(A\cup B)  + m_{2}(B)m_1(A\cup B) \\
&= (3/6)\cdot (1/6) +(3/6)\cdot (1/6) + (1/6)\cdot (2/6) = 8/36
\end{align*}
\begin{align*}
m_{\wedge}(A\cup B) & = m_{1}(A\cup B)m_2(A\cup B) = (2/6)\cdot (1/6)= 2/36
\end{align*}
%
\begin{align*}
m_{\wedge}(A\cap B) &  \triangleq m_{\wedge}(A\cap B) = m_{1}(A)m_2(B) + m_{2}(B)m_1(B)\\
& = (1/6)\cdot (1/6) + (4/6)\cdot (3/6) = 13/36
\end{align*}


\noindent
$\bullet$ {\bf{Dempster's rule} \cite{Dempster_1968}}: 

\noindent This combination rule has been initially proposed by Dempster and then used by Shafer in DST framework. We assume (without loss of generality) that the sources of evidence are equally reliable. Otherwise a discounting preprocessing is first applied. It is defined on $G^\Theta=2^\Theta$ by forcing $m_{DS}(\emptyset)\triangleq 0$ and $\forall A \in G^\Theta\setminus\{\emptyset\}$ by:
\begin{equation}
m_{DS}(A) = \frac{1}{1-k} m_{\wedge}(A)=\frac{ m_{\wedge}(A)}{1-m_{\wedge}(\emptyset)}.
\label{eq:DSR}
 \end{equation}
When $k=1$, this rule cannot be used. Dempster's rule of combination can be directly extended for the combination of $N$ independent and equally reliable sources of evidence and its major interest comes essentially from its commutativity and associativity properties. Dempster's rule corresponds to the normalized conjunctive rule by reassigning the mass of total conflict onto all focal elements through the conjunctive operator. The problem enlightened by the famous Zadeh's example \cite{Zadeh_1979} is the repartition of the global conflict. Indeed, consider $\Theta=\{A,B,C\}$ and two experts opinions given by $m_1(A)=0.9$, $m_1(C)=0.1$, and $m_2(B)=0.9$, $m_2(C)=0.1$, the mass given by Dempster's combination is $m_{DS}(C)=1$ which looks very counter-intuitive since it reflects the minority opinion. The generalized Zadeh's example proposed by Smarandache and Dezert in \cite{DSmTBook1}, shows that the results obtained by Dempster's rule can moreover become totally independent of the numerical values taken by $m_1(.)$ and $m_2(.)$ which is much more surprising and difficult to accept without reserve for practical fusion applications. To resolve this problem, Smets \cite{Smets90b} suggested in his Transferable Belief Model (TBM) framework \cite{Smets94} to consider $\Theta$ as an open-world and therefore to use the conjunctive rule instead Dempster's rule at the credal level. At credal level $m_{\wedge}(\emptyset)$  is interpreted as a non-expected solution. The problem is actually just postponed by Smets at the decision/pignistic level since the normalization (division by $1-m_{\wedge}(\emptyset)$) is also required in order to compute the pignistic probabilities of elements of $\Theta$. In other words, the non normalized version of Dempster's rule corresponds to the Smets' fusion rule in the TBM framework working under an open-world assumption, \emph{i.e.} $m_S(\emptyset)=k=m_{\wedge}(\emptyset)$ and $\forall A \in G^\Theta\setminus\{\emptyset\}$, $m_S(A)=m_{\wedge}(A)$.\\

\noindent
{\it{Example 2}}: Let's consider the 2D frame and quantitative masses as given in example 1 and assume Shafer's model (\emph{i.e.} $A\cap B=\emptyset$), then the conflicting quantitative mass $k=m_{\wedge}(A\cap B)=13/36$ is redistributed to the sets $A$, $B$, $A\cup B$ proportionally with their $m_{\wedge}(.)$ masses, \emph{i.e.} $m_{\wedge}(A)=13/36$, $m_{\wedge}(B)=8/36$ and $m_{\wedge}(A\cup B)=2/36$ respectively through Demspter's rule \eqref{eq:DSR}. One thus gets:
\begin{align*}
m_{DS}(\emptyset) &  = 0\\
m_{DS}(A) &  = (13/36)/ (1 - (13/36)) = 13/23\\
m_{DS}(B) &  = (8/36)/ (1 - (13/36)) = 8/23\\
m_{DS}(A\cup B) &  = (2/36)/ (1 - (13/36)) = 2/23
\end{align*}
\noindent
If one prefers to adopt Smets' TBM approach, at the credal level the empty set is now allowed to have positive mass. In this case, one gets:
\begin{align*}
m_{TBM}(\emptyset) &  = m_{\wedge}(A\cap B)= 13/36\\
m_{TBM}(A) &  = 13/36\\
m_{TBM}(B) &  = 8/36\\
m_{TBM}(A\cup B) &  = 2/36
\end{align*}

\noindent
$\bullet$ {\bf{Yager's rule}} \cite{Yager_1983, Yager_1985,Yager87}: 

\noindent Yager admits that in case of high conflict Dempster's rule provides counter-intuitive results. Thus, $k$ plays the role of an absolute discounting term added to the weight of ignorance. The commutative and quasi-associative\footnote{quasi-associativity was defined by Yager in \cite{Yager87}, and Smarandache and Dezert in \cite{Smarandache06}.} Yager's rule is given by $m_Y(\emptyset)=0$ and $\forall A \in G^\Theta\setminus\{\emptyset\}$ by
\begin{equation}
\begin{cases}
m_Y(A)=m_{\wedge}(A) \\
m_Y(\Theta)=m_{\wedge}(\Theta) + m_{\wedge}(\emptyset)
\end{cases}
\label{eq:YagerRule}
\end{equation}


\noindent
{\it{Example 3}}: Let's consider the 2D frame and quantitative masses as given in example 1 and assume Shafer's model (\emph{i.e.} $A\cap B=\emptyset$), then the conflicting quantitative mass $k=m_{\wedge}(A\cap B)=13/36$ is transferred to total ignorance $A\cup B$. One thus gets:
\begin{align*}
m_{Y}(A) &  = 13/36\\
m_{Y}(B) &  = 8/36\\
m_{Y}(A\cup B) &  = (2/36) + (13/36)= 15/36
\end{align*}

\noindent
$\bullet$ {\bf{Dubois \& Prade's rule}} \cite{Dubois88}: 

\noindent This rule supposes that the two sources are reliable when they are not in conflict and at least one of them is right when a conflict occurs. Then if one believes that a value is in a set $X$ while the other believes that this value is in a set $Y$, the truth lies in $X\cap Y$ as long $X\cap Y\neq \emptyset$. If $X\cap Y=\emptyset$, then the truth lies in $X\cup Y$. According to this principle, the commutative and quasi-associative Dubois \& Prade hybrid rule of combination, which is a reasonable trade-off between precision and reliability, is defined by $m_{DP}(\emptyset)=0$ and $\forall A \in G^\Theta\setminus\{\emptyset\}$ by
\begin{equation}
m_{DP}(A)=m_{\wedge}(A)  + \sum_{\substack{X,Y\in G^\Theta\\ X\cup Y=A \\X\cap Y=\emptyset}} m_1(X)m_2(Y)
\label{eq:DuboisRule}
\end{equation}
\noindent
In Dubois \& Prade's rule, the conflicting information is considered more precisely than in Dempster's or Yager's rules since all partial conflicts involved the total conflict are taken into account separately through \eqref{eq:DuboisRule}.\\

The repartition of the conflict is very important because of the non-idempotency of the rules (except the Den\oe ux' rule \cite{Denoeux06} that can be applied when the dependency between experts is high) and due to the responses of the experts that can be conflicting. Hence, we have defined the auto-conflict \cite{Osswald06} in order to quantify the intrinsic conflict of a mass and the distribution of the conflict according to the number of experts.\\


\noindent
{\it{Example 4}}: Taking back example 1 and assuming Shafer's model for $\Theta$,  the quantitative Dubois \& Prade's rule gives the same result as quantitative Yager's rule since the conflicting mass, $m_{\wedge}(A\cap B)=13/36$, is transferred to $A\cup B$, while the other quantitative masses remain unchanged.\\

$\bullet$ {\bf{Proportional Conflict Redistribution rules (PCR)}}:

\subsection*{$\PCRdsm$ for combining two sources}

\noindent
Smarandache and Dezert proposed five proportional conflict redistribution ($\PCR$) methods \cite{Smarandache05,Smarandache06} to redistribute the partial conflict on the elements implied in the partial conflict. The most efficient for combining two basic belief assignments $m_1(.)$ and $m_2(.)$ is the $\PCRdsm$ rule
given by $m_\PCRdsm(\emptyset)=0$ and for all $X \in G^\Theta$, $X\neq \emptyset$ by:

\begin{equation}
\label{DSmTcombination}
m_\PCRdsm(X)=m_\wedge(X) +\sum_{\substack{Y\in G^\Theta\\ X\cap Y \equiv \emptyset}} 
\left(\frac{m_1(X)^2 m_2(Y)}{m_1(X) \!+\! m_2(Y)}+\frac{m_2(X)^2 m_1(Y)}{m_2(X) \!+\! m_1(Y)}\!\right) \, ,
 \end{equation}

\noindent
where $m_\wedge(.)$ is the conjunctive rule given by the equation \eqref{eq:comb_conj}. \\

\noindent
{\it{Example 5}}: Let's consider the 2D frame and quantitative masses as given in example 1 and assume Shafer's model (\emph{i.e.} $A\cap B=\emptyset$), then the conflicting quantitative mass $k=m_{\wedge}(A\cap B)=13/36$ is redistributed only to elements involved in conflict, $A$ and $B$ (not to $A\cup B$). We repeat that:
\begin{align*}
m_{\wedge}(A\cap B) & = m_{1}(A)m_2(B) + m_{2}(B)m_1(B) = (1/6)\cdot (1/6) + (4/6)\cdot (3/6) = 13/36.
\end{align*}
\noindent
So $(1/6)\cdot (1/6)=1/36$ is redistributed to $A$ and $B$ proportionally to their quantitative masses assigned by the sources (or experts) $m_1(A)=1/6$ and $m_2(B)=1/6$:
$$\frac{x_{1,A}}{1/6}=\frac{y_{1,B}}{1/6}=\frac{1/36}{(1/6)+(1/6)}=1/12,$$
\noindent
hence
$$x_{1,A}=(1/6) \cdot (1/12)= 1/72,$$
\noindent
and
$$y_{1,B}=(1/6) \cdot (1/12)= 1/72.$$
\noindent
Similarly $(4/6)\cdot (3/6) = 12/36$ is redistributed to $A$ and $B$ proportionally to their quantitative masses assigned by the sources (or experts) $m_2(A)=4/6$ and $m_1(B)=3/6$:
$$\frac{x_{2,A}}{4/6}=\frac{y_{2,B}}{3/6}=\frac{12/36}{(4/6)+(3/6)}=2/7,$$
\noindent
hence
$$x_{2,A}=(4/6) \cdot (2/7)=4/21,$$
\noindent
and
$$y_{2,B}=(3/6) \cdot (2/7) = 1/7.$$
\noindent
It is easy to check that 
$$x_{1,A}+ y_{1,B} + x_{2,A}+ y_{2,B}=13/36=m_{\wedge}(A\cap B)$$

\noindent
Summing, we get:
\begin{align*}
& m_{\PCRdsm}(A) = (13/36) + (1/72) + (4/21) = 285/504 \simeq 0.57\\
& m_{\PCRdsm}(B) = (8/36) + (1/72) + (1/7) =191/504\simeq 0.38\\
& m_{\PCRdsm}(A\cup B) = 2/36 \simeq 0.05\\
& m_{\PCRdsm}(A\cap B = \emptyset) = 0
\end{align*}

\noindent
\subsection*{$\PCRmo$ for combining more than two sources}

A generalization of $\PCRdsm$ fusion rule for combining altogether more than two experts has been proposed by Smarandache and Dezert in \cite{Smarandache06}. Recently Martin and Osswald \cite{Martin06b, Osswald06} studied and formulated a new version of the $\PCRdsm$ rule, denoted $\PCRmo$,  for combining more than two sources, say $M$ sources with $M\geq 2$. Martin and Osswald have shown that $\PCRmo$ exhibits a better behavior than $\PCRdsm$
 in specific interesting cases. $\PCRmo$ rule is defined as follows: $m_\PCRmo(\emptyset)=0$  and for all $X \in G^\Theta$, $X\neq \emptyset$,
\begin{eqnarray}
\label{PCR6combination}
\begin{array}{c}
  \displaystyle m_\PCRmo(X)  =  \displaystyle m_\wedge(X) + \sum_{i=1}^M
  m_i(X)^2 \displaystyle \sum_{\begin{array}{c}
      \scriptstyle {\displaystyle \mathop{\cap}_{k=1}^{M\!-\!1}} Y_{\sigma_i(k)} \cap X \equiv \emptyset \\
      \scriptstyle (Y_{\sigma_i(1)},...,Y_{\sigma_i(M\!-\!1)})\in (G^\Theta)^{M\!-\!1}
  \end{array}}
  \!\!\!\!\!\!\!\!\!\!\!\!
  \left(\!\!\frac{\displaystyle \prod_{j=1}^{M\!-\!1} m_{\sigma_i(j)}(Y_{\sigma_i(j)})}
       {\displaystyle m_i(X) \!+\! \sum_{j=1}^{M\!-\!1} m_{\sigma_i(j)}(Y_{\sigma_i(j)})}\!\!\right)\!\!,
\end{array}
\end{eqnarray}
where $Y_j \in G^\Theta$ is the response of the expert $j$, $m_j(Y_j)$ the associated belief function and $\sigma_i$ counts from 1 to $M$ avoiding $i$:
\begin{eqnarray}
\label{sigma}
\left\{
\begin{array}{ll}
\sigma_i(j)=j &\mbox{if~} j<i,\\
\sigma_i(j)=j+1 &\mbox{if~} j\geq i.\\
\end{array}
\right.
\end{eqnarray}

The idea is here to redistribute the masses of the focal elements giving a partial conflict proportionally to the initial masses on these elements. \\

In general, for $M\geq 3$ sources, one calculates the total conflict, which is a sum of products; if each product is formed by factors of masses of distinct hypothesis, then $\PCRmo$ coincides with $\PCRdsm$; if at least a product is formed by at least two factors of masses of same hypotheses, then $\PCRmo$ is different from $\PCRdsm$:
\begin{itemize}
\item for example: a product like $m_1(A)m_2(A)m_3(B)$, herein we have two masses of hypothesis $A$;
\item or $m_1(A\cup B)m_2(B\cup C)m_3(B\cup C)m_4(B\cup C)$, herein we have three masses of hypothesis $B\cup C$ from four sources.\\
\end{itemize}

\noindent
{\it{Example 6}}: For instance, consider three experts expressing their opinion on $\Theta=\{A,B,C,D\}$ in the Shafer's model:
\begin{table}[h]
\centering
  \begin{tabular}{|l|c|c|c|c|}
    \hline
    & $A$ & $B$ & $A\cup C$ & $A\cup B \cup C \cup D$ \\
    \hline
    $m_1(.)$  & 0.7 & 0 & 0 & 0.3 \\
    \hline
    $m_2(.)$ & 0 & 0.5 & 0 & 0.5 \\
    \hline
    $m_3(.)$ & 0 & 0 & 0.6 & 0.4 \\
    \hline
  \end{tabular}
    \caption{Quantitative inputs for example 6}
\label{TableExample6}
\end{table}

The global conflict is given here by 0.21+0.14+0.09=0.44, coming from:
\begin{itemize}
\item[-]  $A$, $B$ and $A\cup C$ for the partial conflict 0.21,
\item[-]  $A$, $B$ and $A\cup B \cup C \cup D$ for 0.14,
\item[-]  and $B$, $A\cup C$ and $A\cup B \cup C \cup D$ for 0.09.\\
\end{itemize}

With the generalized $\PCRmo$ rule (\ref{PCR6combination}), we obtain:
\begin{align*}
m_\PCRmo(A) & = 0.14+0.21+0.21\cdot\frac{7}{18}+0.14 \cdot\frac{7}{16} \simeq 0.493\\
m_\PCRmo(B)& = 0.06+0.21\cdot\frac{5}{18} +0.14\cdot\frac{5}{16}+0.09\cdot\frac{5}{14} \simeq 0.194\\
m_\PCRmo(A\cup C) &= 0.09+0.21\cdot\frac{6}{18}+0.09\cdot\frac{6}{14}  \simeq 0.199\\
m_\PCRmo(A\cup B \cup C \cup D) & = 0.06+0.14\cdot\frac{4}{16}+0.09\cdot\frac{3}{14} \simeq 0.114
\end{align*}

\noindent
{\it{Example 7}}: Let's consider three sources providing quantitative belief masses only on unions.
\begin{table}[h]
\centering
  \begin{tabular}{|l|c|c|c|c|}
    \hline
    & $A\cup B$ & $B\cup C$ & $A\cup C$ & $A\cup B\cup C$ \\
    \hline
    $m_1(.)$  & 0.7 & 0 & 0 & 0.3 \\
    \hline
    $m_2(.)$ & 0 & 0 & 0.6 & 0.4 \\
    \hline
    $m_3(.)$ & 0 & 0.5 & 0 & 0.5 \\
    \hline
  \end{tabular}
    \caption{Quantitative inputs for example 7}
\label{TableExample7}
\end{table}

\noindent
The conflict is given here by:
%
\begin{align*}
m_{\wedge}(\emptyset)  & = m_1(A\cup B)m_2(A \cup  C)m_3(B \cup C)  = 0.7\cdot 0.6 \cdot0.5=0.21
\end{align*}

\noindent
With the generalized PCR rule, \emph{i.e.} PCR6, we obtain:
$$m_\PCRmo(A) = 0.21,$$
$$m_\PCRmo(B) = 0.14,$$
$$m_\PCRmo(C) = 0.09,$$
$$m_\PCRmo(A\cup B) = 0.14+0.21.\frac{7}{18}\simeq 0.2217,$$
$$m_\PCRmo(B \cup C) = 0.06+0.21.\frac{5}{18} \simeq 0.1183,$$
$$m_\PCRmo(A \cup C) = 0.09+0.21.\frac{6}{18} = 0.16,$$
$$m_\PCRmo(A\cup B\cup C) = 0.06.$$

\noindent
In the sequel, we use the notation $\PCR$ for two and more sources.


\section{Classical qualitative combination rules}
\label{classical qualitative combination_rules}
\label{Sec4}

The classical qualitative combination rules are direct extensions of classical quantitative rules presented in previous section. Since the formulas of qualitative fusion rules are the same as for quantitative rules, they will be not reported in this section. The main difference between quantitative and qualitative approaches lies in the addition, multiplication and division operators one has to use. For quantitative fusion rules, one uses addition, multiplication and division operators on numbers while for  qualitative fusion rules one uses the addition, multiplication and division operators on linguistic labels defined as in section \ref{QualitativeOperators}. \\

\noindent
{\it{Example 8}}: Below is a very simple example used to show how classical qualitative fusion rules work. Let's consider the following set of linguistic labels $L=\{L_{\min}=L_0,L_1,L_2,L_3,L_4,L_5,L_{\max}=L_6\}$ and let's assume Shafer's model for the frame $\Theta=\{A,B\}$ we want to work on. In this example, we consider only two experts providing the qualitative belief assignments (masses) $qm_1(.)$ and $qm_2(.)$ such that:

\begin{table}[h]
\centering
  \begin{tabular}{|l|c|c|c|}
    \hline
    & $A$ & $B$ & $A\cup B$ \\
    \hline
    $qm_1(.)$  & $L_1$ & $L_3$ & $L_2$ \\
    \hline
   $qm_2(.)$  & $L_4$ & $L_1$ & $L_1$ \\
     \hline
  \end{tabular}
    \caption{Qualitative inputs for example 8}
\label{TableExample8}
\end{table}

\noindent The qualitative belief assignments $qm_1(.)$ and $qm_2(.)$ have been chosen quasi-normalized since $L_1+L_3+L_2=L_6=L_{\max}$ and respectively $L_4+L_1+L_1=L_6=L_{\max}$.\\

\noindent
$\bullet$ {\bf{Qualitative Conjunctive Rule}} (QCR): This rule provides $qm_{\wedge}(.)$ according following derivations:
%
\begin{align*}
qm_{\wedge}(A) & = qm_{1}(A)qm_2(A) + qm_{1}(A)qm_2(A\cup B)  + qm_{2}(A)qm_1(A\cup B) \\
&= L_1 L_4 + L_1 L_1 + L_4 L _2 = L_{\frac{1\cdot 4}{6}} + L_{\frac{1\cdot 1}{6}} + L_{\frac{4\cdot 2}{6}} = L_{\frac{4+1+8}{6}} = L_{\frac{13}{6}}\\
qm_{\wedge}(B) & = qm_{1}(B)qm_2(B) + qm_{1}(B)qm_2(A\cup B)  + qm_{2}(B)qm_1(A\cup B) \\
&= L_3 L_1 + L_3 L_1 + L_1 L _2 = L_{\frac{3\cdot 1}{6}} + L_{\frac{3\cdot 1}{6}} + L_{\frac{1\cdot 2}{6}} = L_{\frac{3+3+2}{6}} = L_{\frac{8}{6}}\\
qm_{\wedge}(A\cup B) & = qm_{1}(A\cup B)qm_2(A\cup B)  = L_2 L_1 = L_{\frac{2\cdot 1}{6}}= L_{\frac{2}{6}}\\
qm_{\wedge}(A\cap B) & = qm_{1}(A)qm_2(B) + qm_{2}(B)qm_1(B) = L_1 L_1 + L_4 L_3  = L_{\frac{1\cdot 1}{6}} + L_{\frac{4\cdot 3}{6}}= L_{\frac{1+12}{6}}=L_{\frac{13}{6}}
\end{align*}
We see that not approximating the indexes (\emph{i.e.} working with refined labels), the quasi-normalization of the qualitative conjunctive rule is kept:
$$ L_{\frac{13}{6}} + L_{\frac{8}{6}} +  L_{\frac{2}{6}} + L_{\frac{13}{6}}=L_{\frac{36}{6}}=L_6=L_{\max}.$$
But if we approximate each refined label, we get:
%
\begin{align*}
L_{[\frac{13}{6}]} + L_{[\frac{8}{6}]} +  L_{[\frac{2}{6}]} + L_{[\frac{13}{6}]}& =L_2 + L_1 + L_0 + L_2=L_5 \neq L_6=L_{\max}.
\end{align*}

Let's examine the transfer of the conflicting qualitative mass $qm_{\wedge}(A\cap B)=qm_{\wedge}(\emptyset)=L_{\frac{13}{6}}$ to the non-empty sets according to the main following combination rules:\\

\noindent
$\bullet$ {\bf{Qualitative Dempster's rule}} (extension of classical numerical DS rule to qualitative masses): 
\noindent
Assuming Shafer's model for the frame $\Theta$ (\emph{i.e.} $A\cap B=\emptyset$) and according to DS rule, the conflicting qualitative mass $qm_{\wedge}(A\cap B)=L_{\frac{13}{6}}$ is redistributed to the sets $A$, $B$, $A\cup B$ proportionally with their $qm_{\wedge}(.)$ masses $L_{\frac{13}{6}}$, $L_{\frac{8}{6}}$, and $L_{\frac{2}{6}}$ respectively:

%
\begin{align*}
\frac{x_A}{L_{\frac{13}{6}}}=\frac{y_B}{L_{\frac{8}{6}}}=\frac{z_{A\cup B}}{L_{\frac{2}{6}}}& =\frac{L_{\frac{13}{6}}}{L_{\frac{13}{6}}+L_{\frac{8}{6}}+L_{\frac{2}{6}}} =\frac{L_{\frac{13}{6}}}{L_{\frac{23}{6}}} 
 = L_{(\frac{13}{6} \div \frac{23}{6})\cdot 6} = L_{(\frac{13}{23})\cdot 6} = L_{\frac{78}{23}}
\end{align*}

\noindent
Therefore, one gets:

$$x_A=  L_{\frac{13}{6}} \cdot L_{\frac{78}{23}} = L_{( \frac{13}{6} \cdot \frac{78}{23}) \div 6} = L_{\frac{169}{138}}$$

$$y_B=  L_{\frac{8}{6}} \cdot L_{\frac{78}{23}} = L_{( \frac{8}{6} \cdot \frac{78}{23}) \div 6} = L_{\frac{104}{138}}$$

$$z_{A\cup B}=  L_{\frac{2}{6}} \cdot L_{\frac{78}{23}} = L_{( \frac{2}{6} \cdot \frac{78}{23}) \div 6} = L_{\frac{26}{138}}$$

\noindent
We can check that the  qualitative conflicting mass, $L_{\frac{13}{6}}$, has been proportionally split into three qualitative masses:
$$L_{\frac{169}{138}} +  L_{\frac{104}{138}} + L_{\frac{26}{138}} = L_{\frac{169+104+26}{138}}=L_{\frac{299}{138}}=L_{\frac{13}{6}}$$
\noindent
So:
\begin{align*}
& qm_{DS}(A) = L_{\frac{13}{6}} +L_{\frac{169}{138}} = L_{\frac{13}{6} + \frac{169}{138}} = L_{\frac{468}{138}}\\
& qm_{DS}(B) = L_{\frac{8}{6}} +L_{\frac{104}{138}} = L_{\frac{8}{6} + \frac{104}{138}} = L_{\frac{288}{138}}\\
& qm_{DS}(A\cup B) = L_{\frac{2}{6}} +L_{\frac{26}{138}} = L_{\frac{2}{6} + \frac{26}{138}} = L_{\frac{72}{138}}\\
& qm_{DS}(A\cap B=\emptyset) = L_{0} 
\end{align*}

\noindent
$qm_{DS}(.)$ is quasi-normalized since:
$$L_{\frac{468}{138}} + L_{\frac{288}{138}} + L_{\frac{72}{138}}=L_{\frac{828}{138}}=L_6=L_{\max}.$$

\noindent
If we approximate the linguistic labels $L_{\frac{468}{138}}$, $L_{\frac{288}{138}}$ and  $L_{\frac{72}{138}}$ in order to work with original labels in $L$, still $qm_{DS}(.)$ remains quasi-normalized since:
\begin{align*}
& qm_{DS}(A) \approx L_{[\frac{468}{138}]}=L_3\\
& qm_{DS}(B) \approx L_{[\frac{288}{138}]}=L_2\\
& qm_{DS}(A\cup B) \approx L_{[\frac{72}{138}]}=L_1
\end{align*}
\noindent
and $L_3+L_2+L_1=L_6=L_{\max}$.\\

\noindent
$\bullet$ {\bf{Qualitative Yager's rule}}: \\

\noindent
With Yager's rule, the qualitative conflicting mass $L_{\frac{13}{6}}$ is entirely transferred to the total ignorance $A\cup B$, so:
$$qm_{Y}(A\cup B) = L_{\frac{2}{6}} + L_{\frac{13}{6}}= L_{\frac{15}{6}}$$
\noindent
and 
\noindent
$$qm_{Y}(A\cap B) = qm_{Y}(\emptyset)=L_0$$
\noindent
while the others remain the same:
$$qm_{Y}(A) =L_{\frac{13}{6}}$$
$$qm_{Y}(B) =L_{\frac{8}{6}}$$

\noindent
$qm_{Y}(.)$ is quasi-normalized since:
$$L_{\frac{13}{6}} + L_{\frac{8}{6}} + L_{\frac{15}{6}}=L_{\frac{36}{6}}=L_6=L_{\max}.$$

\noindent
If we approximate the linguistic labels $L_{\frac{13}{6}}$, $L_{\frac{8}{6}}$ and  $L_{\frac{15}{6}}$, still $qm_{Y}(.)$ happens to remain quasi-normalized since:
\begin{align*}
& qm_{Y}(A) \approx L_{[\frac{13}{6}]}=L_2\\
& qm_{Y}(B) \approx L_{[\frac{8}{6}]}=L_1\\
& qm_{Y}(A\cup B) \approx L_{[\frac{15}{6}]}=L_3
\end{align*}
\noindent
and $L_2+L_1+L_3=L_6=L_{\max}$.\\

\noindent
$\bullet$ {\bf{Qualitative Dubois \& Prade's rule}}: In this example the Qualitative Dubois \& Prade's rule gives the same result as qualitative Yager's rule since the conflicting mass, $qm_{\wedge}(A\cap B)=L_{\frac{13}{6}}$, is transferred to $A\cup B$, while the other qualitative masses remain unchanged.\\

\noindent
$\bullet$ {\bf{Qualitative Smets' TBM rule}}:  Smets' TBM approach allows keeping mass on the empty set. One gets:
\begin{align*}
& qm_{TBM}(A) = L_{\frac{13}{6}}\\
& qm_{TBM}(B) = L_{\frac{8}{6}}\\
& qm_{TBM}(A\cup B) = L_{\frac{2}{6}}\\
& qm_{TBM}(\emptyset) = L_{\frac{13}{6}}
\end{align*}
\noindent
Of course $qm_{TBM}(.)$ is also quasi-normalized.\\

\noindent
However if we approximate, $qm_{TBM}(.)$ does not remain quasi-normalized in this case since:
\begin{align*}
& qm_{TBM}(A) \approx L_{[\frac{13}{6}]}=L_2\\
& qm_{TBM}(B) \approx L_{[\frac{8}{6}]}=L_1\\
& qm_{TBM}(A\cup B) \approx L_{[\frac{2}{6}]}=L_0\\
& qm_{TBM}(\emptyset) \approx L_{[\frac{13}{6}]}=L_2
\end{align*}
\noindent
and $L_2+L_1+L_0+L_2=L_5\neq L_6=L_{\max}$.\\

\noindent
$\bullet$ {\bf{Qualitative PCR (QPCR)}}: The conflicting qualitative mass, $qm_{\wedge}(A\cap B)=L_{\frac{13}{6}}$, is redistributed only to elements involved in conflict, $A$ and $B$ (not to $A\cup B$). We repeat that
\begin{align*}
qm_{\PCR}(A\cap B) & = qm_{1}(A)qm_2(B) + qm_{2}(B)qm_1(B)\\
& = L_1 L_1 + L_4 L_3 = L_{\frac{1\cdot 1}{6}} + L_{\frac{4\cdot 3}{6}}= L_{\frac{1+12}{6}}=L_{\frac{13}{6}}
\end{align*}

\noindent
So $L_{\frac{1}{6}}$ is redistributed to $A$ and $B$ proportionally to their qualitative masses assigned by the sources (or experts) $qm_1(A)=L_1$ and $qm_2(B)=L_1$:
$$\frac{x_{1,A}}{L_1}=\frac{y_{1,B}}{L_1}=\frac{L_{\frac{1}{6}}}{L_1+L_1}=\frac{L_{\frac{1}{6}}}{L_2}=L_{(\frac{1}{6}\div 2)\cdot 6}=L_{\frac{1}{2}}$$
\noindent
hence
$$x_{1,A}=L_1\cdot L_{\frac{1}{2}}= L_{(1\cdot \frac{1}{2})\div 6}=L_{\frac{1}{12}}$$
\noindent
and
$$y_{1,B}=L_1\cdot L_{\frac{1}{2}}= L_{\frac{1}{12}}$$
\noindent
Similarly $L_{\frac{12}{6}}$ is redistributed to $A$ and $B$ proportionally to their qualitative masses assigned by the sources (or experts) $qm_2(A)=L_4$ and $qm_1(B)=L_3$:
$$\frac{x_{2,A}}{L_4}=\frac{y_{2,B}}{L_3}=\frac{L_{\frac{12}{6}}}{L_4+L_3}=\frac{L_{\frac{12}{6}}}{L_7}=L_{(\frac{12}{6}\div 7)\cdot 6}=L_{\frac{12}{7}}$$
\noindent
hence
$$x_{2,A}=L_4\cdot L_{\frac{12}{7}}= L_{(4\cdot \frac{12}{7})\div 6}=L_{\frac{8}{7}}$$
\noindent
and
$$y_{2,B}=L_3\cdot L_{\frac{12}{7}}= L_{(3\cdot \frac{12}{7})\div 6}=L_{\frac{6}{7}}$$
\noindent
Summing, we get:
\begin{align*}
& qm_{\PCR}(A) = L_{\frac{13}{6}} + L_{\frac{1}{12}} + L_{\frac{8}{7}}= L_{\frac{285}{84}}\\
& qm_{\PCR}(B) = L_{\frac{8}{6}} + L_{\frac{1}{12}} + L_{\frac{6}{7}}= L_{\frac{191}{84}}\\
& qm_{\PCR}(A\cup B) = L_{\frac{2}{6}}=L_{\frac{28}{84}}\\
& qm_{\PCR}(A\cap B = \emptyset) = L_{0}
\end{align*}

\noindent
$qm_{\PCR}(.)$ is quasi-normalized since:

$$L_{\frac{285}{84}} + L_{\frac{191}{84}}+ L_{\frac{28}{84}}=L_{\frac{504}{84}}=L_6=L_{\max}.$$

\noindent
However, if we approximate, it is not quasi-normalized any longer since:
$$L_{[\frac{285}{84}]} + L_{[\frac{191}{84}]}+ L_{[\frac{28}{84}]}=
L_3 + L_2 + L_0=L_5 \neq L_6=L_{\max}.$$

In general, if we do not approximate, and we work with quasi-normalized qualitative masses, no matter what fusion rule we apply, the result will be quasi-normalized. If we approximate, many times the quasi-normalization is lost.

\section{Generalization of quantitative fusion rules}
\label{general_formulation}
\label{Sec5}

In \cite{Smets97, Appriou05} we can find two propositions of a general formulation of the combination rules. In the first one, Smets considers the combination rules from a matrix notation and find the shape of this matrix according to some assumptions on the rule, such as linearity, commutativity, associativity, etc. In the second one, a generic operator is defined from the plausibility functions. 

A general formulation of the global conflict repartition have been proposed in \cite{Inagaki91, Lefevre02a} for all $X\in 2^\Theta$ by:
\begin{eqnarray}
\label{rep}
m_c(X)=m_{\wedge}(X)+w(X)m_{\wedge}(\emptyset),
\end{eqnarray}
where $\displaystyle \sum_{X\in 2^\Theta} w(X)=1$. The problem is the choice of the weights $w(X)$.\\

\subsection{How to choose conjunctive and disjunctive rules?}
\label{Mix_conj_dis}

We have seen that conjunctive rule reduces the imprecision and uncertainty but can be used only if one of the experts is reliable, whereas the disjunctive rule can be used when the experts are not reliable, but allows a loss of specificity. 

Hence, Florea \cite{Florea06} proposes a weighted sum of these two rules according to the global conflict $k=m_{\wedge}(\emptyset)$ given for $X \in 2^\Theta$ by:
\begin{eqnarray}
\label{Florea}
m_\Flo(X)=\displaystyle \beta_1(k) m_{\vee}(X)+ \beta_2(k) m_{\wedge}(X),
\end{eqnarray}
where $\beta_1$ and $\beta_2$ can admit $\displaystyle k=\frac{1}{2}$ as symmetric weight: 
\begin{eqnarray}
\begin{array}{l}
\beta_1(k)=\displaystyle \frac{k}{1-k+k^2},\\
\beta_2(k) =\displaystyle \frac{1-k}{1-k+k^2}.\\
\end{array}
\end{eqnarray}
Consequently, if the global conflict is high ($k$ near 1) the behavior of this rule will give more importance to the disjunctive rule. Thus, this rule considers the global conflict coming from the non-reliability of the experts. 

In order to take into account the weights more precisely in each partial combination, we propose the following new rule. For two basic belief assignments $m_1$ and $m_2$ and for all $X \in G^\Theta$, $X\neq \emptyset$ we have:

%
\begin{equation}
m_\mix(X)=\sum_{Y_1\cup Y_2 =X} \delta_1(Y_1,Y_2) m_1(Y_1)m_2(Y_2) +  \sum_{Y_1 \cap Y_2 =X} \delta_2(Y_1,Y_2) m_1(Y_1)m_2(Y_2).
\label{MixQuantitative}
\end{equation}

Of course, if $ \delta_1(Y_1,Y_2)=\beta_1(k)$ and $\delta_2(Y_1,Y_2)=\beta_2(k)$ we obtain Florea's rule. In the same manner, if $\delta_1(Y_1,Y_2)=1- \delta_2(Y_1,Y_2)=0$ we obtain the conjunctive rule and if $\delta_1(Y_1,Y_2)=1- \delta_2(Y_1,Y_2)=1$ the disjunctive rule. If $\delta_1(Y_1,Y_2)=1- \delta_2(Y_1,Y_2)=\ind_{Y_1\cap Y_2 =\emptyset}$ we retrieve Dubois and Prade's rule and the partial conflict can be considered, whereas the rule (\ref{Florea}).\\

The choice of $\delta_1(Y_1,Y_2)=1- \delta_2(Y_1,Y_2)$ can be done by a dissimilarity such as: 
\begin{eqnarray}
\label{eq_delta}
\delta_1(Y_1,Y_2)= \delta(Y_1,Y_2)\triangleq \displaystyle 1-\frac{\mathcal{C}(Y_1 \cap Y_2)}{\min \{\mathcal{C}(Y_1), \mathcal{C}(Y_2)\} },
\end{eqnarray}
or
\begin{eqnarray}
\label{eq_eta}
\delta_1(Y_1,Y_2)= \eta(Y_1,Y_2)\triangleq \displaystyle 1-\frac{\mathcal{C}(Y_1 \cap Y_2)}{\max \{\mathcal{C}(Y_1), \mathcal{C}(Y_2)\} },
\end{eqnarray}
where $\mathcal{C}(Y_1)$ is the cardinality of $Y_1$. In the case of the DST framework,  $\mathcal{C}(Y_1)$ is the number of distinct elements of $Y_1$. In the case of the DSmT,  $\mathcal{C}(Y_1)$ is the DSm cardinality given by the number of parts of $Y_1$ in the Venn diagram of the problem \cite{DSmTBook1}. 
$\delta(.,.)$ in \eqref{eq_delta} is actually not a proper dissimilarity measure (\emph{e.g.} $\delta(Y_1,Y_2)=0$ does not imply $Y_1=Y_2$), but $\eta(.,.)$ defined in \eqref{eq_eta} is a proper dissimilarity measure. We can also take for $\delta_2(Y_1,Y_2)$, the Jaccard's distance, \emph{i.e.} $\delta_2(Y_1,Y_2)=d(Y_1,Y_2)$ given by:
\begin{eqnarray}
\label{eq_d2}
d(Y_1,Y_2)=\displaystyle \frac{\mathcal{C}(Y_1 \cap Y_2)}{\mathcal{C}(Y_1 \cup Y_2)},
\end{eqnarray}
used by \cite{Jousselme01} on the belief functions. Note that $d$ is not a distance in the case of DSmT. Thus, if we have a partial conflict between $Y_1$ and $Y_2$, $\mathcal{C}(Y_1 \cap Y_2)=0$ and the rule transfers the mass on $Y_1 \cup Y_2$. In the case $Y_1\subset Y_2$ (or the contrary), $Y_1 \cap Y_2=Y_1$ and $Y_1 \cup Y_2=Y_2$, so with $\delta_1(.,.)=\delta(.,.)$ the rule transfers the mass on $Y_1$ and with $\delta_1(.,.)=1-d(.,.)$  it transfers the mass on $Y_1$ and $Y_2$ according to the ratio ($\mathcal{C}(Y_1)/\mathcal{C}(Y_2)$) of the cardinalities. In the case  $Y_1 \cap Y_2 \neq Y_1, Y_2$ and $\emptyset$, the rule transfers the mass on $Y_1 \cap Y_2$ and $Y_1 \cup Y_2$ according to $\delta(.,.)$ and $d(.,.)$.\\

\noindent
{\it{Example 9}}: (on the derivation of the weights)\\

Let's consider a frame of discernment $\Theta=\{A,B,C\}$ in Shafer's model (\emph{i.e.} all intersections empty).

\begin{enumerate}

\item[a)] We compute the first similarity weights $\delta_2(.,.)=1-\delta(.,.)$:
\begin{table}[h]
\centering
  \begin{tabular}{|c|c|c|c|c|}
    \hline
   $\delta_2(.,.)=1-\delta(.,.)$ & $A$ & $B$ & $C$ & $A\cup B$ \\
    \hline
    $A$  & 1 & 0 & 0 & 1 \\
    \hline
   $B$  & 0 & 1 & 0 & 1 \\
    \hline
   $C$  & 0 & 0 & 1 & 0 \\
    \hline
   $A\cup B$   & 1 & 1 & 0 & 1 \\
     \hline
  \end{tabular}
\caption{Values for $1-\delta(.,.)$}
\label{TableEx9a}
\end{table}
\noindent
since
$$\delta_2(A,A)= \frac{\mathcal{C}(A\cap A)}{\min \{\mathcal{C}(A), \mathcal{C}(A)\} }= \frac{\mathcal{C}(A)}{\mathcal{C}(A)}=1$$
$$\delta_2(A,B)= \frac{\mathcal{C}(A\cap B)}{\min \{\mathcal{C}(A), \mathcal{C}(B)\} }= 0$$
\noindent
because $A\cap B=\emptyset$ and $\mathcal{C}(\emptyset)=0$.
$$\delta_2(A,A\cup B)= \frac{\mathcal{C}(A\cap (A\cup B))}{\min \{\mathcal{C}(A), \mathcal{C}(A\cup B)\} }= \frac{\mathcal{C}(A)}{\mathcal{C}(A)}=1$$
\noindent
etc.\\

\noindent
Whence, the first dissimilarity weights $\delta_1(.,.)$ defined by \eqref{eq_delta}, \emph{i.e.} $\delta_1(X,Y)=1- \delta_2(X,Y)$ take the values:
\begin{table}[h]
\centering
  \begin{tabular}{|c|c|c|c|c|}
    \hline
   $\delta_1(.,.)=\delta(.,.)$ & $A$ & $B$ & $C$ & $A\cup B$ \\
    \hline
    $A$  & 0 & 1 & 1 & 0 \\
    \hline
   $B$  & 1 & 0 & 1& 0 \\
    \hline
   $C$  & 1 & 1 & 0 & 1 \\
    \hline
   $A\cup B$   & 0 & 0 & 1 & 0 \\
     \hline
  \end{tabular}
\caption{Values for $\delta(.,.)$}
\label{TableEx9b}
\end{table}

The first similarity and dissimilarity weights $\delta_2(.,.)$ and $\delta_1(.,.)$ are not quite accurate, since for example: $\delta_2(A,A\cup B)=1$, \emph{i.e.} $A$ and $A\cup B$ are 100\% similar (which is not the case since $A\neq A\cup B$) and $\delta_1(A,A\cup B)=1-\delta_2(A,A\cup B)=1-1=0$, \emph{i.e.} $A$ and $A\cup B$ are 100\% dissimilar (which is not the case either since $A\cap (A\cup B)\neq \emptyset$).\\

\item[b)] The second similarity weights $\delta_2(.,.)=1-\eta(.,.)$ given by the equation (\ref{eq_eta}) overcomes this problem. We obtain on the previous example:
\begin{table}[h]
\centering
  \begin{tabular}{|c|c|c|c|c|}
    \hline
   $\delta_2(.,.)=1-\eta(.,.)$ & $A$ & $B$ & $C$ & $A\cup B$ \\
    \hline
    $A$  & 1 & 0 & 0 & 1/2 \\
    \hline
   $B$  & 0 & 1 & 0 & 1/2 \\
    \hline
   $C$  & 0 & 0 & 1 & 0 \\
    \hline
   $A\cup B$   & 1/2 & 1/2 & 0 & 1 \\
     \hline
  \end{tabular}
\caption{Values for $1-\eta(.,.)$}
\label{TableEx9c}
\end{table}
\noindent
since
$$\delta_2(A,A)=1-\eta(A,A)= \frac{\mathcal{C}(A\cap A)}{\max \{\mathcal{C}(A), \mathcal{C}(A)\} }= \frac{\mathcal{C}(A)}{\mathcal{C}(A)}=1$$
$$\delta_2(A,B)=1-\eta(A,B)= \frac{\mathcal{C}(A\cap B)}{\max \{\mathcal{C}(A), \mathcal{C}(B)\} }= 0$$
\noindent
because $A\cap B=\emptyset$ and $\mathcal{C}(\emptyset)=0$.
%
\begin{align*}
\delta_2(A,A\cup B)=1-\eta(A,A\cup B)&= \frac{\mathcal{C}(A\cap (A\cup B))}{\max \{\mathcal{C}(A), \mathcal{C}(A\cup B)\} }= \frac{\mathcal{C}(A)}{\mathcal{C}(A\cup B)}=\frac{1}{2}
\end{align*}
\noindent
which is better than $\delta_2(A,A\cup B)=1-\delta(A,A\cup B)=1$.\\
\noindent
etc.\\

\noindent
Whence, the second dissimilarity weights $\eta(.,.)$ take the values:

\begin{table}[h]
\centering
  \begin{tabular}{|c|c|c|c|c|}
    \hline
   $\delta_1(.,.)=\eta(.,.)$ & $A$ & $B$ & $C$ & $A\cup B$ \\
    \hline
    $A$  & 0 & 1 & 1 & 1/2 \\
    \hline
   $B$  & 1 & 0 & 1& 1/2 \\
    \hline
   $C$  & 1 & 1 & 0 & 1 \\
    \hline
   $A\cup B$   & 1/2 & 1/2 & 1 & 0 \\
     \hline
  \end{tabular}
\caption{Values for $\eta(.,.)$}
\label{TableEx9d}
\end{table}

\noindent
$\eta(A,A\cup B)=1-\frac{1}{2}=\frac{1}{2}$, which is better than $\delta_1(A,A\cup B)=\delta(A,A\cup B)=0$.\\

The second similarity weight coincides with Jaccard's distance in Shafer's model, but in hybrid and free models, they are generally different. Hence if we consider a Shafer's model, one gets for all $Y_1$, $Y_2$ in $G^\Theta$:
$$d(Y_1,Y_2)=1-\eta(Y_1,Y_2).$$
Smarandache defined in \cite{Smarandache06InDepth} the degree of intersection of two sets as Jaccard's distance, and also the degree of union of two sets, and the degree of inclusion of a set into another set and improved many fusion rules by inserting these degrees in the fusion rules' formulas. \\
\end{enumerate}

\noindent
{\it{Example 10}}: (with Shafer's model)\\

\noindent
Consider the following example given in the table~\ref{TableEx10} for two (quantitative) experts providing $m_1(.)$ and $m_2(.)$ on $\Theta=\{A,B,C\}$ and let's assume that Shafer's model holds (\emph{i.e.} $A$, $B$ and $C$ are truly exclusive):
\begin{table}[h]
\centering
  \begin{tabular}{|c|c|c|c|c|c|}
    \hline
    & $m_1(.)$ & $m_2(.)$ & $m_{\wedge}$ & $m_{\mix, \delta}$  & $m_{\mix, \eta}$  \\
    &  &  &  &   & $m_{\mix, d}$  \\
    \hline
    $\emptyset$  & 0& 0 & 0.2  & 0 & 0 \\
    \hline
     $A$ & 0.3 & 0 & 0.3 & 0.24 & 0.115  \\
    \hline
    $B$  & 0 & 0.2 & 0.14  & 0.14 & 0.06 \\
    \hline
    $A\cup B$ & 0.4 & 0 & 0.12 & 0.18 & 0.18 \\
    \hline
    $C$ & 0 & 0.2 & 0.06 & 0.06 & 0.02 \\
    \hline
    $A\cup C$ & 0 & 0.3& 0.09 & 0.15& 0.165\\
    \hline
    $A\cup B \cup C$ & 0.3 & 0.3 & 0.09 & 0.23 & 0.46\\
    \hline
  \end{tabular}
\caption{Quantitative inputs and fusion result}
\label{TableEx10}
\end{table}

\noindent
When taking $\delta_{1}(.,.)=\delta(.,.)$ according to \eqref{eq_delta}, one obtains:
\begin{table}[h]
\centering
  \begin{tabular}{|c|c|c|c|}
    \hline
    $\delta_1(.,.)=\delta(.,.)$ &  $A$ &  $A\cup B$ & $A\cup B \cup C$ \\
    \hline
    $B$  & 1& 0 & 0  \\
    \hline
    $C$  & 1& 1 & 0  \\
    \hline
    $A\cup C$ & 0 & 1/2 & 0 \\
    \hline
    $A\cup B \cup C$ & 0 & 0 & 0 \\
    \hline
  \end{tabular}
\caption{Values for $\delta(.,.)$}
\label{TableEx10b}
\end{table}

\noindent
where the columns are the focal elements of the basic belief assignment given by the expert 1 and the rows are the focal elements of the basic belief assignment given by expert 2. The mass 0.2 on $\emptyset$ come from the responses $A$ and $C$ with a value of 0.06, from the responses $A$ and $B$ with a value of 0.06 and from the responses $A \cup B$ and $C$ with a value of 0.08. These three values are transferred respectively on $A\cup C$, $A\cup B$ and $A\cup B\cup C$. The mass 0.12 on $A$ given by the responses $A\cup B$ and $A\cup C$ is transferred on $A$ with a value of 0.06 and on $A\cup B\cup C$ with the same value. \\

\noindent
When taking $\delta_{1}(.,.)=\eta(.,.)$ or $\delta_{1}(.,.)=1-d(.,.)$ according to \eqref{eq_eta} and \eqref{eq_d2}, one obtains:
\begin{table}[h]
\centering
  \begin{tabular}{|c|c|c|c|}
    \hline
    $\delta_1(.,.)=\eta(.,.)$ & & &  \\
    $\delta_1(.,.)=1-d(.,.)$ &  $A$ &  $A\cup B$ & $A\cup B \cup C$ \\
    \hline
    $B$  & 1& 1/2 & 2/3  \\
    \hline
    $C$  & 1& 1 & 2/3  \\
    \hline
    $A\cup C$ & 1/2 & 2/3 & 1/3 \\
    \hline
    $A\cup B \cup C$ & 2/3 & 1/3 & 0 \\
    \hline
  \end{tabular}
\caption{Values for $\eta(.,.)$ or $1-d(.,.)$}
\label{TableEx10c}
\end{table}

\noindent
With $\delta_1(.,.)=\eta$ or $\delta_1(.,.)=1-d(.,.)$, the rule is more disjunctive: more masses are transferred on the ignorance.\\

Note that $\delta_1(.,.)=\delta(.,.)$ can be used when the experts are considered reliable. In  this case,  we consider the most precise response. With $\delta_1(.,.)=\eta(.,.)$ or $\delta_1(.,.)=1-d(.,.)$, we get the conjunctive rule only when the experts provide the same response, otherwise we consider the doubtful responses and we transfer the masses in proportion of the imprecision of the responses (given by the cardinality of the responses) on the part in agreement and on the partial ignorance. \\

\noindent
{\it{Example 11}}: (with a hybrid model)\\

\noindent
Consider the same example with two (quantitative) experts providing $m_1(.)$ and $m_2(.)$ on the frame of discernment $\Theta=\{A,B,C\}$ with the following integrity constraints: $A\cap B \neq \emptyset$, $A\cap C=\emptyset$ and $B\cap C= \emptyset$ (which defines a so-called DSm-hybrid model \cite{DSmTBook1}):
\begin{table}[h]
\centering
  \begin{tabular}{|c|c|c|c|c|c|c|}
    \hline
    & $m_1(.)$ & $m_2(.)$ & $m_{\wedge}$ & $m_{\mix, \delta}$  & $m_{\mix, \eta}$ & $m_{\mix, d}$  \\
    \hline
    $\emptyset$  & 0& 0 & 0.14  & 0 & 0 & 0 \\
    \hline
     $A \cap B$ & 0 & 0 & 0.06 & 0.03 & 0.03 & 0.02  \\
    \hline
     $A$ & 0.3 & 0 & 0.3 & 0.26 & 0.205 & 0.185  \\
    \hline
    $B$  & 0 & 0.2 & 0.14  & 0.14 & 0.084 & 0.084 \\
    \hline
    $A\cup B$ & 0.4 & 0 & 0.12 & 0.15 & 0.146 & 0.156 \\
    \hline
    $C$ & 0 & 0.2 & 0.06 & 0.06 & 0.015 & 0.015 \\
    \hline
    $A\cup C$ & 0 & 0.3& 0.09 & 0.15 & 0.1575 & 0.1575\\
    \hline
    $A\cup B \cup C$ & 0.3 & 0.3 & 0.09 & 0.21 & 0.3625 & 0.3825\\
    \hline
  \end{tabular}
\caption{Quantitative inputs and fusion result}
\label{TableEx11}
\end{table}


\noindent
When taking $\delta_{1}(.,.)=\delta(.,.)$ according to \eqref{eq_delta}, one obtains:
\begin{table}[h]
\centering
  \begin{tabular}{|c|c|c|c|}
    \hline
    $\delta_{1}(.,.)=\delta(.,.)$ &  $A$ &  $A\cup B$ & $A\cup B \cup C$ \\
    \hline
    $B$  & 1/2& 0 & 0  \\
    \hline
    $C$  & 1& 1 & 0  \\
    \hline
    $A\cup C$ & 0 & 1/3 & 0 \\
    \hline
    $A\cup B \cup C$ & 0 & 0 & 0 \\
    \hline
  \end{tabular}
\caption{Values for $\delta(.,.)$}
\label{TableEx11b}
\end{table}

\noindent
When taking $\delta_{1}(.,.)=\eta(.,.)$ according to \eqref{eq_eta}, one obtains:
\begin{table}[h]
\centering
  \begin{tabular}{|c|c|c|c|}
    \hline
    $\delta_{1}(.,.)=\eta(.,.)$ &  $A$ &  $A\cup B$ & $A\cup B \cup C$ \\
    \hline
    $B$  & 1/2& 1/3 & 1/2  \\
    \hline
    $C$  & 1& 1 & 3/4  \\
    \hline
    $A\cup C$ & 1/3 & 1/3 & 1/4 \\
    \hline
    $A\cup B \cup C$ & 1/2 & 1/4 & 0 \\
    \hline
  \end{tabular}
\caption{Values for $\eta(.,.)$}
\label{TableEx11c}
\end{table}

\noindent
When taking $\delta_{1}(.,.)=1-d(.,.)$ according to \eqref{eq_d2}, one obtains:
\begin{table}[h]
\centering
  \begin{tabular}{|c|c|c|c|}
    \hline
    $\delta_1(.,.)=1-d(.,.)$ &  $A$ &  $A\cup B$ & $A\cup B \cup C$ \\
    \hline
    $B$  & 2/3& 1/3 & 1/2  \\
    \hline
    $C$  & 1& 1 & 3/4  \\
    \hline
    $A\cup C$ & 1/3 & 1/2 & 1/4 \\
    \hline
    $A\cup B \cup C$ & 1/2 & 1/4 & 0 \\
    \hline
 \end{tabular}
\caption{Values for $1-d(.,.)$}
\label{TableEx11d}
\end{table}

For more than two experts, say $M>2$, if the intersection of the responses of the $M$ experts is not empty, we can still transfer on the intersection and the union, and the equations (\ref{eq_delta}) and (\ref{eq_eta}) become:
\begin{eqnarray}
\label{delta}
\delta_1(Y_1,...,Y_M)=\delta(Y_1,...,Y_M)=\displaystyle 1-\frac{\mathcal{C}(Y_1 \cap ... \cap Y_M)}{\displaystyle \min_{1\leq i \leq M}\mathcal{C}(Y_i)},
\end{eqnarray} 
and
\begin{eqnarray}
\label{eta}
\delta_1(Y_1,...,Y_M)=\eta(Y_1,...,Y_M)=\displaystyle 1-\frac{\mathcal{C}(Y_1 \cap ... \cap Y_M)}{\displaystyle \max_{1\leq i \leq M}\mathcal{C}(Y_i)}.
\end{eqnarray}
From equation (\ref{eq_d2}), we can define $\delta_1$ by:
\begin{eqnarray}
\label{d}
\delta_1(Y_1,...,Y_M)=\displaystyle 1-\frac{\mathcal{C}(Y_1 \cap ... \cap Y_M)}{\mathcal{C}(Y_1 \cup ... \cup Y_M)}.
\end{eqnarray} 

Finally, the mixed rule for $M\geq 2$ experts is given by:
%
\begin{equation}
\label{mixed}
m_\mix(X)=\sum_{Y_1 \cup ... \cup Y_M=X} \delta_1(Y_1,...,Y_M) \prod_{j=1}^M m_j(Y_j)
 +  \sum_{Y_1 \cap ... \cap Y_M=X} (1-\delta_1(Y_1,...,Y_M)) \prod_{j=1}^M m_j(Y_j).\\
\end{equation}

This formulation can be interesting according to the coherence of the responses of the experts. However, it does not allow the repartition of the partial conflict in an other way than the Dubois and Prade's rule. 

\subsection{A discounting proportional conflict repartition rule}
\label{general_PCR}

The $\PCRmo$ redistributes the masses of the conflicting focal elements proportionally to the initial masses on these elements. 
First, the repartition concerns only on the elements involved in the partial conflict. We can apply a discounting procedure in the combination rule in order to transfer a part of the partial conflict on the partial ignorance. This new discounting $\PCR$ (noted $\DPCR$) can be expressed for two basic belief assignments $m_1(.)$ and $m_2(.)$ and for all $X \in G^\Theta$, $X\neq \emptyset$ by:


%
\begin{align}
m_\DPCR(X)=m_{\wedge}(X) & + 
\sum_{\substack{Y\in G^\Theta\\ X\cap Y \equiv \emptyset}} \alpha \cdot \left(\frac{m_1(X)^2 m_2(Y)}{m_1(X) + m_2(Y)}+\frac{m_2(X)^2 m_1(Y)}{m_2(X) \!+\! m_1(Y)}\!\right)\nonumber \\
& + \sum_{\substack{Y_1\cup Y_2 =X\\Y_1\cap Y_2 \equiv \emptyset }}
(1-\alpha)  \cdot m_1(Y_1)m_2(Y_2),
\label{DPCRrule}
\end{align}

\noindent
where $\alpha\in [0,1]$ is the discounting factor. Note that we can also apply a discounting procedure on the masses before the combination as shown in \eqref{massDisounted}. Here the discounting factor is introduced in order to transfer a part of the partial conflict on partial ignorance. We propose in \eqref{alpha} and \eqref{g_function} different ways for choosing this factor $\alpha$.\\

Hence, $\DPCR$ fusion rule is a combination of $\PCR$ and Dubois-Prade (or DSmH\footnote{The DSmH rule is an extension of Dubois-Prade's rule which has been proposed in the DSmT framework in order to work with hybrid models including non-existential constraints. See \cite{DSmTBook1} for details and examples.})
rules. In an analogue way we can combine other fusion rules, two or more in the same formula, getting new mixed formulas. So that in a general case, for $M\geq 2$ experts, we can extend the previous rule as:

%
\begin{align}
\label{DPCR}  
 m_\DPCR(X) = m_{\wedge}(X) & + \sum_{i=1}^M m_i(X)^2 \!\!\!\!\!\!\!\!\!\!\!\! 
 \sum_{\substack{ \displaystyle \mathop{\cap}_{k=1}^{M-1} Y_{\sigma_i(k)} \cap X = \emptyset \\  \displaystyle (Y_{\sigma_i(1)},...,Y_{\sigma_i(M-1)})\in (G^\Theta)^{M-1}}}
 \!\!\!\!\!\!\!\!\!\!\!\! \alpha \cdot \left(\!\!\frac{\displaystyle \prod_{j=1}^{M-1} m_{\sigma_i(j)}(Y_{\sigma_i(j)})}
{\displaystyle m_i(X) + \sum_{j=1}^{M-1} m_{\sigma_i(j)}(Y_{\sigma_i(j)})}\!\!\right)\nonumber\\
&+ \sum_{\substack{Y_1 \cup ... \cup Y_M = X\\ Y_1 \cap ... \cap Y_M \equiv \emptyset}} 
(1-\alpha) \cdot \prod_{j=1}^M m_j(Y_j),
\end{align}

where $Y_j \in G^\Theta$ is a response of the expert $j$, $m_j(Y_j)$ its assigned mass and $\sigma_i$ is given by (\ref{sigma}).\\

Hence, if we choose as discounting factor $\alpha=0.9$ in the previous example, we obtain:
\begin{align*}
m_\DPCR(A) & = 0.14+0.21+0.21 \cdot\frac{7}{18}\cdot 0.9+0.14\cdot\frac{7}{16}\cdot 0.9 \simeq 0.479\\
m_\DPCR(B) & = 0.06+0.21\cdot\frac{5}{18}\cdot 0.9+0.14\cdot\frac{5}{16}\cdot  0.9 +0.09\cdot \frac{5}{14} \cdot 0.9 \simeq 0.181\\
m_\DPCR(A\cup C) & = 0.09+0.21\cdot\frac{6}{18}\cdot 0.9+0.09\cdot \frac{6}{14}\cdot 0.9 \simeq 0.187\\
m_\DPCR(A\cup B \cup C)& = 0.21 \cdot  0.1 = 0.021\\
m_\DPCR(A\cup B \cup C \cup D) & = 0.06+0.14\cdot\frac{4}{16}\cdot 0.9 +0.09\cdot\frac{3}{14}\cdot 0.9+ 0.14 \cdot  0.1 + 0.09 \cdot  0.1\simeq 0.132
\end{align*}

However, in this example, the partial conflict due to the experts 1, 2 and 3 saying $A$, $B$, and $A \cup C$ respectively, the conflict is 0.21. Nonetheless, only the experts 1 and 2 and the experts 2 and 3 are in conflict. The experts 1 and 3 are not in conflict. \\

Now, consider another case where the experts 1, 2 and 3 say $A$, $B$, and $C$ respectively with the same conflict 0.21. In both cases, the $\DPCR$ rule transfers the masses with the same weight $\alpha$. Although, we could prefer transfer more mass on $\Theta$ in the second than in the first case.\\ 

Consequently, the transfer of mass can depend on the existence of conflict between each pair of experts. We define the conflict function giving the number of experts in conflict two by two for each response $Y_i \in G^\Theta$ of the expert $i$ as the number of responses of the other experts in conflict with $i$. A function $f_i$ is defined by the mapping of $(G^\Theta)^M$ onto $\left[0,\displaystyle \frac{1}{M}\right]$ with:
\begin{eqnarray}
\label{f_function}
f_i(Y_1,...,Y_M)=\displaystyle \frac{\displaystyle  \sum_{j=1}^M\ind_{\{Y_j\cap Y_i= \emptyset\}}}{M(M-1)}.
\end{eqnarray}

Hence, we can choose $\alpha$ depending on the response of the experts such as:
\begin{eqnarray}
\label{alpha}
\alpha(Y_1,...,Y_M)=1-\sum_{i=1}^M f_i(Y_1,...,Y_M).
\end{eqnarray}
In this case $\alpha \in [0,1]$, we do not transfer the mass on elements that can be written as the union of the responses of the experts.\\

Therefore, if we consider again our previous example we obtain:
$$\alpha(A,B, A\cup C)=1-\frac{2}{3}=\frac{1}{3},$$
$$ \alpha(A,B, A\cup B \cup C \cup D)=1-\frac{1}{3}=\frac{2}{3},$$
$$\alpha(A\cup B \cup C \cup D,B, A\cup C)=1-\frac{1}{3}=\frac{2}{3}.$$
Thus the provided mass by the $\DPCR$ is:
\begin{align*}
m_\DPCR(A)& = 0.14+0.21+0.21 \cdot \frac{7}{18}\cdot  \frac{1}{3}+0.14\cdot  \frac{7}{16} \cdot\frac{2}{3} \simeq 0.418\\
m_\DPCR(B) & = 0.06+0.21  \cdot \frac{5}{18} \cdot \frac{1}{3}+0.14  \cdot\frac{5}{16}  \cdot \frac{2}{3} +0.09  \cdot \frac{5}{14}  \cdot\frac{2}{3} \simeq 0.130\\
m_\DPCR(A\cup C) & = 0.09+0.21 \cdot \frac{6}{18}\cdot\frac{1}{3}+0.09 \cdot\frac{6}{14} \cdot\frac{2}{3}\simeq 0.139\\
m_\DPCR(A\cup B \cup C)& = 0.21 \cdot \frac{2}{3} = 0.140\\
m_\DPCR(A\cup B \cup C\cup D) & = 0.06+0.14 \cdot \frac{4}{16} \cdot \frac{2}{3} +0.09 \cdot \frac{3}{14} \cdot \frac{2}{3} + 0.14 \cdot \frac{1}{3}   + 0.09 \cdot \frac{1}{3} \simeq 0.173
\end{align*}

We want to take account of the degree of conflict (or non-conflict) within each pair of expert differently for each element. We can consider the non-conflict function given for each expert $i$ by the number of experts not in conflict with $i$. Hence, we can choose $\alpha_i(Y_1,...,Y_M)$ defined by the mapping of $(G^\Theta)^M$ onto $\left[\displaystyle 0, \frac{1}{M} \right]$ with:
%
\begin{equation}
\label{g_function}
\alpha_i(Y_1,...,Y_M)= \displaystyle \frac{1}{M}- f_i(Y_1,...,Y_M)=\displaystyle \frac{ \displaystyle  \sum_{j=1, j\neq i}^M\ind_{\{Y_j\cap Y_i \not\equiv \emptyset\}}}{M (M-1)}.
\end{equation}

The discounting PCR rule (equation (\ref{DPCR})) can be written for $M$ experts, for all $X \in G^\Theta$, $X\neq \emptyset$ as:

%
\begin{align}
\label{DPCRi}
m_\DPCR(X)  =  m_{\wedge}(X) & + \sum_{i=1}^M m_i(X)^2 
\sum_{\substack{ \displaystyle \mathop{\cap}_{k=1}^{M-1} Y_{\sigma_i(k)} \cap X = \emptyset \\  \displaystyle (Y_{\sigma_i(1)},...,Y_{\sigma_i(M-1)})\in (G^\Theta)^{M-1}}}
  \alpha_i \lambda \left(\!\!\frac{\displaystyle \prod_{j=1}^{M\!-\!1} m_{\sigma_i(j)}(Y_{\sigma_i(j)})}
       {\displaystyle m_i(X) \!+\! \sum_{j=1}^{M\!-\!1} m_{\sigma_i(j)}(Y_{\sigma_i(j)})}\!\!\right)\nonumber\\
& + \sum_{\substack{Y_1 \cup ... \cup Y_M = X \\ Y_1 \cap ... \cap Y_M \equiv \emptyset}}
 (1-\sum_{i=1}^M \alpha_i) \prod_{j=1}^M m_j(Y_j),
\end{align}

\noindent
where $\alpha_i(X,Y_{\sigma_i(1)},...,Y_{\sigma_i(M-1)})$ is noted $\alpha_i$ for notations convenience and $\lambda$ depending on $(X,Y_{\sigma_i(1)},...,Y_{\sigma_i(M-1)})$, is chosen to obtain the normalization given by the equation (\ref{normDST}). $\lambda$ is given when $\alpha_i\neq 0$, $\forall i \in \{1,...,M\}$ by:

\begin{eqnarray}
\lambda=\frac{\displaystyle \sum_{i=1}^M \alpha_i}{<\boldsymbol{\alpha},\boldsymbol{\gamma}>},
\end{eqnarray}

\noindent
where $<\boldsymbol{\alpha},\boldsymbol{\gamma}>$ is the scalar product of $\boldsymbol{\alpha}=(\alpha_i)_{i\in \{1,...,M\}}$ and $\boldsymbol{\gamma}=(\gamma_i)_{i\in \{1,...,M\}}$ with:

\begin{eqnarray}
\gamma_i = \frac{\displaystyle m_i(X)}
       {\displaystyle m_i(X) \!+\! \sum_{j=1}^{M\!-\!1} m_{\sigma_i(j)}(Y_{\sigma_i(j)})},
\end{eqnarray}

\noindent
where $\gamma_i(X,Y_{\sigma_i(1)},...,Y_{\sigma_i(M-1)})$ is noted $\gamma_i$ for notations convenience.\\

With this last version of the rule, for $\alpha_i$ given by the equation (\ref{g_function}), we obtain on our illustrative example  $\lambda=\frac{36}{13}$ when the experts 1, 2 and 3 say $A$, $B$, and $A \cup C$ respectively (the conflict is 0.21), $\lambda=\frac{16}{5}$ when the conflict is 0.14 and $\lambda=\frac{56}{17}$ when the conflict is 0.09. Thus, the masses are given by:
\begin{align*}
m_\DPCR(A) & = 0.14+0.21+0.21 \cdot \frac{7}{18}  \cdot \frac{1}{6}  \cdot  \frac{36}{13} +0.14  \cdot  \frac{7}{16}  \cdot  \frac{1}{6}  \cdot  \frac{16}{5} \simeq 0.420\\
m_\DPCR(B) & = 0.06+ 0.14  \cdot  \frac{5}{16}  \cdot  \frac{1}{6}  \cdot  \frac{16}{5} +0.09  \cdot  \frac{5}{14}  \cdot  \frac{1}{6}  \cdot  \frac{56}{17} \simeq 0.101\\
 m_\DPCR(A\cup C) & = 0.09+0.21  \cdot  \frac{6}{18}  \cdot  \frac{1}{6}  \cdot  \frac{36}{13} +0.09  \cdot  \frac{6}{14}  \cdot  \frac{1}{6}  \cdot  \frac{56}{17} \simeq 0.143\\
 m_\DPCR(A\cup B \cup C)& = 0.21  \cdot  \frac{2}{3} =0.14\\
 m_\DPCR(A\cup B \cup C\cup D) & = 0.06+0.14  \cdot  \frac{4}{16}  \cdot  \frac{1}{3}  \cdot  \frac{16}{5} +0.09  \cdot  \frac{3}{14}  \cdot \frac{1}{3}  \cdot  \frac{56}{17}  + 0.14  \cdot  \frac{1}{3}+ 0.09  \cdot  \frac{1}{3} \simeq 0.196
\end{align*}

This last rule of combination allows one to consider a ``kind of degree'' of conflict (a degree of pair of non-conflict), but this degree is not so easy to introduce in the combination rule.

\subsection{A mixed discounting conflict repartition rule}
\label{General_rule}

In this section, we propose a combination of  the mixed rule (\ref{mixed}) with the discounting $\PCR$ (\ref{DPCR}).  This new {\it{mixed discounting conflict repartition rule}} (MDPCR for short) for two quantitative basic belief assignments $m_1(.)$ and $m_2(.)$  is defined by $m_\MDPCR(\emptyset)=0$ and for all $X \in G^\Theta$, $X\neq \emptyset$ by:




\begin{align}
m_\MDPCR(X) = & 
 \sum_{\substack{Y_1\cup Y_2 =X, \\ Y_1\cap Y_2 \not\equiv \emptyset}}
\delta_1(Y_1,Y_2) \cdot m_1(Y_1)m_2(Y_2) \nonumber \\
 & + \sum_{\substack{Y_1\cap Y_2 =X, \\ Y_1\cap Y_2 \not\equiv \emptyset}}
(1-\delta_1(Y_1,Y_2)) \cdot m_1(Y_1)m_2(Y_2)\nonumber \\
& + \sum_{\substack{Y\in G^\Theta, \\ X\cap Y \equiv \emptyset }}
\alpha \cdot \left(\frac{m_1(X)^2 m_2(Y)}{m_1(X) + m_2(Y)}+\frac{m_2(X)^2 m_1(Y)}{m_2(X) + m_1(Y)}\!\right) \nonumber \\
& + \displaystyle\sum_{\substack{Y_1\cup Y_2 =X, \\ Y_1\cap Y_2 \equiv \emptyset}}
(1-\alpha) \cdot m_1(Y_1)m_2(Y_2).
\label{MixedDPCRrule}
\end{align}

$\alpha$ can be given by the equation (\ref{alpha}) and $\delta_1(.,.)$ by the equation (\ref{delta}) or (\ref{d}). The weights must be taken in order to get a kind of continuity between the mixed and DPCR rules. In actuality, when the intersection of the responses is almost empty (but not empty) we use the mixed rule, and when this intersection is empty we chose the $\DPCR$ rule. In the first case, all the mass is transferred on the union, and in the second case it will be the same according to the partial conflict. Indeed, $\alpha=0$ if the intersection is not empty and $\delta_1=1$ if the intersection is empty. We can also introduce $\alpha_i$ given by the equation \eqref{g_function}, and this continuity is conserved.\\

This rule is given in a general case for $M$ experts, by $m_\MDPCR(\emptyset)=0$ and for all $X \in G^\Theta$, $X\neq \emptyset$ by:





\begin{align}
\label{eq_mixed_to_dpcr}
m_\MDPCR(X)=& \displaystyle  \sum_{\substack{Y_1 \cup ... \cup Y_M=X, \\ Y_1 \cap ... \cap Y_M \not\equiv \emptyset }}
\delta_1(Y_1,\ldots,Y_M)\cdot  \prod_{j=1}^M m_j(Y_j)\nonumber\\
& +\displaystyle  \sum_{\substack{Y_1 \cap ... \cap Y_M =X, \\
\scriptstyle Y_1 \cap ... \cap Y_M \not\equiv \emptyset}} 
(1- \delta_1(Y_1,\ldots,Y_M)) \cdot \prod_{j=1}^M m_j(Y_j)\nonumber\\
& +
\displaystyle
\sum_{i=1}^M m_i(X)^2 \displaystyle 
\sum_{\substack{ \displaystyle \mathop{\cap}_{k=1}^{M-1} Y_{\sigma_i(k)} \cap X = \emptyset \\  \displaystyle (Y_{\sigma_i(1)},...,Y_{\sigma_i(M-1)})\in (G^\Theta)^{M-1}}}
  \alpha \cdot \left(\!\!\frac{\displaystyle \prod_{j=1}^{M\!-\!1} m_{\sigma_i(j)}(Y_{\sigma_i(j)})}
       {\displaystyle m_i(X) \!+\! \sum_{j=1}^{M\!-\!1} m_{\sigma_i(j)}(Y_{\sigma_i(j)})}\!\!\right)\nonumber\\
& +
\displaystyle
\sum_{\substack{ Y_1 \cup ... \cup Y_M = X, \\
\scriptstyle Y_1 \cap ... \cap Y_M \equiv \emptyset}}
 (1-\alpha) \cdot \prod_{j=1}^M m_j(Y_j),
\end{align}

\noindent
where $Y_j \in G^\Theta$ is the response of the expert $j$, $m_j(Y_j)$ the associated belief function and $\sigma_i$ is given by (\ref{sigma}). This formula could seem difficult to understand, but it can be implemented easily as shown in \cite{Martin07}. \\

If we take again the previous example, with $\delta_1(.,.)$ given by equation (\ref{delta}), there is no difference with the $\DPCR$. If $\delta_1(.,.)$ is calculated by equation (\ref{d}), the only difference pertains to the mass 0.09 coming from the responses of the three experts: $A\cup B \cup C\cup D$, $A\cup B \cup C\cup D$ and $A \cup C$. This mass is transferred on $A\cup C$ (0.06) and on $A\cup B \cup C\cup D$ (0.03).\\

The rules presented in the previous section, propose a repartition of the masses giving a partial conflict only (when at most two experts are in discord) and do not take heed of the level of imprecision of the responses of the experts (the non-specificity of the responses). The imprecision of the responses of each expert is only considered by the mixed and $\MDPCR$ rules when there is no conflict between the experts. To try to overcome these problems Martin and Osswald have proposed a begin of solutions toward a more general rule \cite{Martin07}.

\section{Generalization of qualitative fusion rules}
\label{qualitativegeneral_formulation}
\label{Sec6}

This section provides two simple examples to show in detail how to extend the generalized quantitative fusion rules proposed in the previous section (\emph{i.e.} the Mixed, the Discounted, and the Mixed Discounted fusion rules) to their qualitative counterparts using our operators on linguistic labels defined in section \ref{QualitativeOperators}.\\

\noindent
{\it{Example 12}}:  Fusion of two sources\\

Consider a set of labels $L=\{L_{\min}=L_0,L_1,L_2,L_3,L_4,L_5,L_{\max}=L_6\}$, and a frame of discernment $\Theta=\{A,B,C\}$ in Shafer's model (\emph{i.e.} all intersections empty). Consider the two following qualitative sources of evidence:

\clearpage
\newpage

\begin{table}[h]
\centering
  \begin{tabular}{|l|c|c|c|c|}
    \hline
    & $A$ & $B$ & $C$ & $A\cup B$ \\
    \hline
    $qm_1(.)$  & $L_2$ & $L_0$ & $L_0$ & $L_4$ \\
    \hline
   $qm_2(.)$  & $L_3$ & $L_2$ & $L_1$ & $L_0$ \\
     \hline
  \end{tabular}
\caption{Qualitative inputs for example 12}
\label{TableEx12}
\end{table}

Now let's apply the qualitative versions of Mixed, Discounted, and Mixed Discounted Quantitative Fusion rules \eqref{MixQuantitative}, \eqref{DPCRrule} and \eqref{MixedDPCRrule} respectively.\\

\noindent
$\bullet$ {\bf{Qualitative Mixed Dubois-Prade's rule}}:
From the formula \eqref{MixQuantitative} and Table \ref{TableEx9b}, one gets:

\begin{align*}
qm_{\mix}^\delta(A)& = \delta(A,A)qm_1(A)qm_2(A) \\
& \qquad + (1-\delta(A,A))qm_1(A)qm_2(A)\\
& \qquad + (1-\delta(A,A\cup B))qm_1(A)qm_2(A\cup B) \\
& \qquad +  (1-\delta(A\cup B,A))qm_1(A\cup B)qm_2(A)\\
& = 0\cdot L_2 L_3 + 1\cdot L_2L_3 + 1\cdot L_2 L_0 + 1\cdot L_4L_3\\
&= L_0 + L_{\frac{2\cdot 3}{6}} + L_{\frac{2\cdot 0}{6}}+ L_{\frac{4\cdot 3}{6}}
=L_{\frac{6}{6}}+ L_{\frac{12}{6}}=L_{\frac{18}{6}}
\end{align*}

\noindent
Similarly, $qm_{\mix}^\delta(B)=L_{\frac{8}{6}}$ and
\begin{align*}
qm_{\mix}^\delta(C)& = \delta(C,C)qm_1(C)qm_2(C)  + (1-\delta(C,C))qm_1(C)qm_2(C)\\
& = 0\cdot L_0 L_1 + 1\cdot L_0L_1 =L_0
\end{align*}
\begin{align*}
qm_{\mix}^\delta(A\cup B)& = \delta(A\cup B,A\cup B)qm_1(A\cup B)qm_2(A\cup B) \\
& \qquad + \delta(A,A\cup B)qm_1(A)qm_2(A\cup B) \\
& \qquad + \delta(A\cup B,A)qm_1(A\cup B)qm_2(A) \\
& \qquad + \delta(B,A\cup B)qm_1(B)qm_2(A\cup B) \\
& \qquad + \delta(A\cup B,B)qm_1(A\cup B)qm_2(B) \\
& \qquad + \delta(A,B)qm_1(A)qm_2(B) + \delta(B,A)qm_1(B)qm_2(A) \\
& \qquad + (1-\delta(A\cup B,A\cup B))qm_1(A\cup B)qm_2(A\cup B) \\
& = L_0 + L_0 + L_0 + L_0 + L_0 + 1\cdot L_2L_2 + 1\cdot L_0L_3\\
& =L_{\frac{2\cdot 2}{6}}+ L_{\frac{0\cdot 3}{6}}=L_{\frac{4}{6}}
\end{align*}

\noindent
Note: The first five terms of previous sum take value $L_0$ since $\delta_1(.,.)=0$ for each of them.
%
\begin{align*}
qm_{\mix}^\delta(A\cup C)& = \delta(A,C)qm_1(A)qm_2(C)  + (1-\delta(C,A))qm_1(C)qm_2(A)\\
& = 1\cdot L_2 L_1 + 1\cdot L_0L_3 = L_{\frac{2\cdot 1}{6}}+ L_{\frac{0\cdot 3}{6}}=L_{\frac{2}{6}}\\
qm_{\mix}^\delta(A\cup B\cup C)& = \delta(C,A\cup B)qm_1(C)qm_2(A\cup B) + \delta(A\cup B,C)qm_1(A\cup B)qm_2(C)\\
& = 1\cdot L_0 L_0 + 1\cdot L_4L_1=L_{\frac{4}{6}}
\end{align*}

This coincides with normal qualitative Dubois-Prade's and DSmH fusion rules. $qm_{\mix}^\delta(.)$ is quasi-normalized both ways:
\begin{itemize}
\item without approximation, since:
\begin{align*}
L_{\frac{18}{6}}+L_{\frac{8}{6}}+L_{\frac{0}{6}}+L_{\frac{4}{6}}+L_{\frac{2}{6}}+L_{\frac{4}{6}} = L_{\frac{18+8+0+4+2+4}{6}}=L_{\frac{36}{6}}=L_{6}=L_{\max}
\end{align*}
\item and with approximations:
\begin{equation*}
L_{[\frac{18}{6}]}+L_{[\frac{8}{6}]}+L_{[\frac{0}{6}]}+L_{[\frac{4}{6}]}+L_{[\frac{2}{6}]}+L_{[\frac{4}{6}]} = L_3+L_1+L_0+L_1+L_0+L_1=L_{6}=L_{\max}
\end{equation*}
\end{itemize}

Compute $qm_{\mix}(.)$ using the second similarity/dissimilarity weights given by the equation \eqref{eq_eta} (which are equal in this case with Jaccard's distance similarity/dissimilarity weights). In this case, we get better results. Since from Table \ref{TableEx9d} and formula \eqref{MixQuantitative}, one gets:
%
%
\begin{align*}
qm_{\mix}^\eta(A)& = \eta(A,A)qm_1(A)qm_2(A)  + (1-\eta(A,A))qm_1(A)qm_2(A)\\
& \qquad + (1-\eta(A,A\cup B))qm_1(A)qm_2(A\cup B)  +  (1-\eta(A\cup B,A))qm_1(A\cup B)qm_2(A)\\
& = 0\cdot L_2 L_3 + 1\cdot L_2L_3 + \frac{1}{2}\cdot L_2 L_0 +  \frac{1}{2}\cdot L_4L_3= L_0 + L_{\frac{2\cdot 3}{6}} + L_{\frac{2\cdot 0}{6\cdot 2}}+ L_{\frac{4\cdot 3}{6\cdot 2}} =L_{0+\frac{6}{6}+ 0+\frac{6}{6}}= L_{\frac{12}{6}}
\end{align*}

\begin{align*}
qm_{\mix}^\eta(B)& = (1-\eta(B,A\cup B))qm_1(B)qm_2(A\cup B)  = \frac{1}{2}\cdot L_2 L_4 =L_{\frac{4\cdot 2}{6\cdot 2}} = L_{\frac{4}{6}}
\end{align*}

%


\begin{align*}
qm_{\mix}^\eta(A\cup B)& = \eta(A\cup B,A\cup B) qm_1(A\cup B)qm_2(A\cup B) \\
& \qquad + \eta(A,A\cup B)qm_1(A)qm_2(A\cup B)  + \eta(A\cup B,A)qm_1(A\cup B)qm_2(A) \\
& \qquad + \eta(B,A\cup B)qm_1(B)qm_2(A\cup B)  + \eta(A\cup B,B)qm_1(A\cup B)qm_2(B) \\
& \qquad + \eta(A,B)qm_1(A)qm_2(B) + \eta(B,A)qm_1(B)qm_2(A) \\
& \qquad + (1-\eta(A\cup B,A\cup B))qm_1(A\cup B)qm_2(A\cup B) \\
& = 0\cdot L_4L_0 +  \frac{1}{2}\cdot L_2L_0 +  \frac{1}{2}\cdot L_4L_3  \frac{1}{2}\cdot L_0L_0 +  \frac{1}{2}\cdot L_4L_2 + 1\cdot L_2L_2 + 1\cdot L_0L_3 + 1\cdot L_4L_0 \\
& = L_0 + L_0 + L_{\frac{4\cdot 3}{6\cdot 2}}+  L_0 + L_{\frac{4\cdot 2}{6\cdot 2}}+ L_{\frac{2\cdot 2}{6}}+ L_0 =L_{\frac{6+4+4}{6}}=L_{\frac{14}{6}}
\end{align*}

\begin{align*}
qm_{\mix}^\eta(A\cup C)& = \eta(A,C)qm_1(A)qm_2(C)  + (1-\eta(C,A))qm_1(C)qm_2(A)\\
& = 1\cdot L_2 L_1 + 1\cdot L_0L_3 = L_{\frac{2\cdot 1}{6}}+ L_{\frac{0\cdot 3}{6}}=L_{\frac{2}{6}}
\end{align*}

\begin{align*}
qm_{\mix}^\eta(A\cup B\cup C)& = \eta(C,A\cup B)qm_1(C)qm_2(A\cup B) + \eta(A\cup B,C)qm_1(A\cup B)qm_2(C)\\
& = 1\cdot L_0 L_0 + 1\cdot L_4L_1=L_{\frac{4}{6}}
\end{align*}

\noindent
Similarly, $qm_{\mix}^\eta(.)$ is quasi-normalized both ways.\\

\noindent
$\bullet$ {\bf{Discounted Qualitative $\PCR$}}: (formula \eqref{DPCRrule}) 

\noindent
We show how to apply the Discounted Qualitative $\PCR$ rule \eqref{DPCRrule} in this example with the fixed discounting factor $\alpha=0.6$, hence $1-\alpha=0.4$. First, apply the qualitative conjunctive rule.

 \begin{table}[h]
\centering
  \begin{tabular}{|l|c|c|c|c|}
    \hline
    & $A$ & $B$ & $C$ & $A\cup B$ \\
    \hline
    $qm_1(.)$  & $L_2$ & $L_0$ & $L_0$ & $L_4$ \\
    \hline
   $qm_2(.)$  & $L_3$ & $L_2$ & $L_1$ & $L_0$ \\
     \hline
  \hline
      $qm_{\wedge}(.)$  & $L_{\frac{18}{6}}$ & $L_{\frac{8}{6}}$ & $L_0$ & $L_0$ \\
     \hline  
  \end{tabular}
  \caption{Qualitative inputs and conjunctive rule}
\label{TableExampleDPCR}
\end{table}

Indeed, one has:
%
\begin{align*}
qm_{\wedge}(A)& =L_2L_3 + L_2L_0 + L_3L_4 = L_{\frac{2\cdot 3}{6} + 0 + \frac{3\cdot 4}{6}}=L_{\frac{18}{6}}\\
qm_{\wedge}(B)& =L_0L_2 + L_0L_0 + L_2L_4 = L_{0+0+\frac{2\cdot 4}{6}}=L_{\frac{8}{6}}\\
qm_{\wedge}(C)& =L_0L_1 =L_0\\
qm_{\wedge}(A\cup B)& =L_4L_0 =L_0
\end{align*}

\noindent
Applying the proportional conflict redistribution according to $\PCR$, one has:

$$\frac{x_{1,A}}{L_2}=\frac{y_{1,B}}{L_2}=\frac{L_2L_2}{L_2+L_2}=\frac{L_{\frac{4}{6}}}{L_4}=L_{(\frac{4}{6}\div 4)\cdot 6}=L_1$$
\noindent
so,
$$x_{1,A}=L_2L_1 = L_{\frac{2}{6}}$$
$$y_{1,B}=L_2L_1 = L_{\frac{2}{6}}$$

$$\frac{x_{2,A}}{L_2}=\frac{z_{1,C}}{L_1}=\frac{L_2L_1}{L_2+L_1}=\frac{L_{\frac{2}{6}}}{L_3}=L_{(\frac{2}{6}\div 3)\cdot 6}=L_{\frac{4}{6}}$$
\noindent
so,
$$x_{2,A}=L_2L_{\frac{4}{6}} = L_{\frac{4/3}{6}}$$
$$z_{1,C}=L_1L_{\frac{4}{6}} = L_{\frac{2/3}{6}}$$

$$\frac{z_{2,C}}{L_1}=\frac{w_{1,A\cup B}}{L_4}=\frac{L_1L_4}{L_1+L_4}=\frac{L_{\frac{4}{6}}}{L_5}=L_{(\frac{4}{6}\div 5)\cdot 6}=L_{\frac{4}{5}}$$
\noindent
so,
$$z_{2,C}=L_1L_{\frac{4}{5}} = L_{\frac{1\cdot 4/5}{6}}= L_{\frac{0.8}{6}}$$
$$w_{1,A\cup B}=L_4L_{\frac{4}{5}} = L_{\frac{4\cdot 4/5}{6}}= L_{\frac{3.2}{6}}$$

\noindent
Summing, we get:
%
\begin{align*}
qm_\DPCR(A)& =L_{\frac{18}{6}} + 0.6 \cdot (L_{\frac{2}{6}} + L_{\frac{4/3}{6}})  = L_{\frac{18}{6}} + 0.6 \cdot L_{\frac{10/3}{6}}  = L_{\frac{18}{6}} + L_{\frac{2}{6}} = L_{\frac{20}{6}}\\
qm_\DPCR(B)& =L_{\frac{8}{6}} + 0.6 \cdot (L_{\frac{2}{6}})  = L_{\frac{8}{6}} + L_{\frac{1.2}{6}}= L_{\frac{9.2}{6}}\\
qm_\DPCR(C)& =L_{0} + 0.6  \cdot (L_{\frac{4/3}{6}} + L_{\frac{0.8}{6}})  = L_0 + L_{\frac{0.6 2/3 + 0.8}{6}} = L_{\frac{0.88}{6}}\\
qm_\DPCR(A\cup B)& =L_{0} + 0.6  \cdot (L_{\frac{3.2}{6}} + 0.4 \cdot (L_2L_2 + L_3L_2) =L_{\frac{1.92}{6}} + 0.4 \cdot  L_{\frac{2\cdot 2}{6}}  = L_{\frac{1.92}{6} + \frac{1.60}{6}}= L_{\frac{3.52}{6}}\\
qm_\DPCR(A\cup C)& =0.4  \cdot (L_2L_1 + L_3L_0) = 0.4  \cdot (L_{\frac{2\cdot 1}{6}} +L_0) = L_{\frac{0.8}{6}}\\
qm_\DPCR(A\cup B \cup C)& =0.4  \cdot (L_0L_0 + L_1L_4)= 0.4  \cdot (L_0 + L_{\frac{1\cdot 4}{6}}) = L_{\frac{1.6}{6}}
\end{align*}

\noindent
We can check that $qm_\DPCR(.)$ is quasi-normalized both ways.\\

\noindent
$\bullet$ {\bf{Mixed Discounted Qualitative $\PCR$}} (formula \eqref{MixedDPCRrule}):\\

In this example, we still set the discounting factor to $\alpha=0.6$.
\begin{enumerate}
\item Using the first kind of similarity/dissimilarity weights (see Table \ref{TableEx9b}), one obtains:
\begin{align*}
qm_{\MDPCR}^{\delta}(A)& = \delta_1(A,A)qm_1(A)qm_2(A)  + \delta_2(A,A)qm_1(A)qm_2(A)\\
& \quad + \delta_2(A,A\cup B)qm_1(A)qm_2(A\cup B)  +  \delta_2(A\cup B,A)qm_1(A\cup B)qm_2(A)\\
& \quad + \alpha  \cdot (L_{\frac{2}{6}} + L_{\frac{4/3}{6}})\\
& = 0 \cdot L_2 L_3 + 1\cdot L_2L_3  + 1\cdot L_2L_0 +1\cdot L_4L_3  + 0.6 \cdot  (L_{\frac{2}{6}} + L_{\frac{4/3}{6}})\\
& = L_{\frac{18}{6}} + 0.6 \cdot  L_{\frac{10/3}{6}} = L_{\frac{18}{6}} + L_{\frac{2}{6}}= L_{\frac{20}{6}}
\end{align*}

\noindent
The term $L_{\frac{4/3}{6}}$ in the sum above comes from the previous Discounted Qualitative $\PCR$ example.\\

\noindent
One gets the same result as in the previous example (Discounted Qualitative $\PCR$).\\

\item Using the second kind of similarity/dissimilarity weights (see Table \ref{TableEx9d}), one obtains:
%
\begin{align*}
qm_{\MDPCR}^{\eta}(A)& = \eta(A,A)qm_1(A)qm_2(A) + (1-\eta(A,A))qm_1(A)qm_2(A)\\
& \quad + (1-\eta(A,A\cup B))qm_1(A)qm_2(A\cup B)  +  (1-\eta(A\cup B,A))qm_1(A\cup B)qm_2(A)\\
& \quad + \alpha  \cdot (L_{\frac{2}{6}} + L_{\frac{4/3}{6}})\\
& = 0 \cdot L_2 L_3 + 1\cdot L_2L_3  + \frac{1}{2}\cdot L_2L_0 + \frac{1}{2}\cdot L_4L_3 + 0.6 \cdot  (L_{\frac{2}{6}} + L_{\frac{4/3}{6}})\\
& = L_{\frac{12}{6}} + L_{\frac{2}{6}}= L_{\frac{14}{6}}
\end{align*}
\noindent
Similarly:
%
%
\begin{align*}
qm_{\MDPCR}^{\eta}(B)& = 0 \cdot L_0 L_2 + 1\cdot L_0L_2  + \frac{1}{2}\cdot L_0L_0 + \frac{1}{2}\cdot L_4L_2  + 0.6 \cdot  L_{\frac{2}{6}} \\
& = \frac{1}{2}\cdot L_4L_2 + L_{\frac{1.2}{6}}= L_{\frac{4\cdot 2}{6\cdot 2}} + L_{\frac{1.2}{6}} =L_{\frac{4}{6}} + L_{\frac{1.2}{6}}= L_{\frac{5.2}{6}}
\end{align*}
\begin{align*}
qm_{\MDPCR}^{\eta}(C)& = 0 \cdot L_0 L_1 + 1\cdot L_0L_1  + 0.6 \cdot  (L_{\frac{2/3}{6}} + L_{\frac{0.8}{6}} ) = L_0 + L_0 + L_{\frac{0.88}{6}}= L_{\frac{0.88}{6}}
\end{align*}

\noindent
The term $L_{\frac{0.8}{6}}$ in the sum above comes from the previous Discounted Qualitative $\PCR$ example.
%
%
\begin{align*}
qm_{\MDPCR}^{\eta}(A\cup B)& = \eta(A\cup B,A\cup B)qm_1(A\cup B)qm_2(A\cup B) \\
& \quad + \eta(A,A\cup B)  qm_1(A)qm_2(A\cup B)  + \eta(A\cup B,A) qm_1(A\cup B)qm_2(A) \\
& \quad + \eta(B,A\cup B) qm_1(B)qm_2(A\cup B)  + \eta(A\cup B,B)qm_1(A\cup B)qm_2(B) \\
& \quad + (1-\eta(A\cup B,A\cup B))qm_1(A\cup B)qm_2(A\cup B) \\
& \quad + \alpha  \cdot L_{\frac{3.2}{6}} + (1-\alpha) qm_1(A)qm_2(B) + (1-\alpha) qm_1(B)qm_2(A) \\
& = 0\cdot L_4L_0 +  \frac{1}{2}\cdot L_0L_1 +  \frac{1}{2}\cdot L_4L_3 \frac{1}{2}\cdot L_0L_0 +  \frac{1}{2}\cdot L_4L_2 + 1\cdot L_4L_0\\
& \quad + 0.6 \cdot L_{\frac{3.2}{6}} + 0.4 \cdot L_2L_2+ 0.4 \cdot L_0L_3\\
& =L_{\frac{4\cdot 3}{6\cdot 2}} + L_{\frac{4\cdot 2}{6\cdot 2}} + L_{\frac{1.92}{6}} + L_{\frac{1.60}{6}} =L_{\frac{6}{6}} + L_{\frac{4}{6}} + L_{\frac{1.92}{6}} + L_{\frac{1.60}{6}} =L_{\frac{13.52}{6}}
\end{align*}

%
\begin{align*}
qm_{\MDPCR}^{\eta}(A\cup C)& = (1-\alpha) qm_1(A)qm_2(C) +  (1-\alpha) qm_1(C)qm_2(A)\\
& = 0.4 \cdot L_2 L_1 + 0.4\cdot L_3L_0=L_{\frac{0.8}{6}}
\end{align*}

%
\begin{align*}
qm_{\MDPCR}^{\eta}(A\cup B\cup C)& = (1-\alpha) qm_1(C)qm_2(A\cup B)  +  (1-\alpha) qm_1(A\cup B)qm_2(C)\\
& = 0.4 \cdot L_0 L_0 + 0.4\cdot L_4L_1=L_{\frac{1.6}{6}}
\end{align*}

\noindent
$qm_{\MDPCR}^{\eta}(.)$ is quasi-normalized without approximations, but it is not with approximations.\\

\end{enumerate}

\clearpage
\newpage

\noindent
{\it{Example 13}}:  Fusion of three sources\\

Consider a set of labels $L=\{L_{\min}=L_0,L_1,L_2,L_3,L_4,L_5,L_{\max}=L_6\}$, and a frame of discernment $\Theta=\{A,B,C\}$ in Shafer's model (\emph{i.e.} all intersections empty). Let's take the three following qualitative sources of evidence:

\begin{table}[h]
\centering
  \begin{tabular}{|l|c|c|c|c|}
    \hline
    & $A$ & $B$ & $B\cup C$ & $A\cup B\cup C$ \\
    \hline
    $qm_1(.)$  & $L_2$ & $L_0$ & $L_0$ & $L_4$ \\
    \hline
   $qm_2(.)$  & $L_0$ & $L_3$ & $L_0$ & $L_3$ \\
     \hline
    $qm_3(.)$  & $L_0$ & $L_0$ & $L_5$ & $L_1$ \\
     \hline
  \end{tabular}
\caption{Qualitative inputs for example 13}
\label{TableEx13}
\end{table}

\noindent
$\bullet$ {\bf{Qualitative Conjunctive Rule}}:\\

If one applies the Qualitative Conjunctive Rule (QCR), one gets:
%
\begin{align*}
qm_{\wedge}(A)& =qm_1(A)qm_2(A\cup B\cup C \cup D)qm_3(A\cup B\cup C \cup D)=L_2L_3L_1=L_{\frac{2\cdot 3}{6}}L_1=L_{\frac{2\cdot 3\cdot 1}{6\cdot 6}}=L_{\frac{1}{6}}
\end{align*}
\noindent
Similarly,
%
%
\begin{align*}
qm_{\wedge}(B)& =L_4L_3L_1 + L_4L_3L_5= L_{\frac{4\cdot 3\cdot 1}{6\cdot 6}}+ L_{\frac{4\cdot 3\cdot 5}{6\cdot 6}}= L_{\frac{2}{6}} + L_{\frac{10}{6}}=L_{\frac{12}{6}}\\
qm_{\wedge}(B\cup C)& =L_4L_3L_5=L_{\frac{4\cdot 3\cdot 5}{6\cdot 6}}=L_{\frac{10}{6}}\\
qm_{\wedge}(A\cup B\cup C \cup D)& =L_4L_3L_1=L_{\frac{4\cdot 3\cdot 1}{6\cdot 6}}=L_{\frac{2}{6}}
\end{align*}

\noindent
The total conflict is:
%
\begin{align*}
qm_{\wedge}(\emptyset)& =qm_1(A)qm_2(B)qm_3(B\cup C) + qm_1(A)qm_2(B)qm_3(A\cup B\cup C \cup D) + qm_1(A)qm_2(A\cup B\cup C \cup D)qm_3(B\cup C)\\
& = L_2L_3L_5 + L_2L_3L_1 + L_2L_3L_5 = L_{\frac{2\cdot 3\cdot 5}{6\cdot 6}} + L_{\frac{2\cdot 3\cdot 1}{6\cdot 6}} +L_{\frac{2\cdot 3\cdot 5}{6\cdot 6}} = L_{\frac{5}{6}} + L_{\frac{1}{6}} + L_{\frac{5}{6}}= L_{\frac{11}{6}}
\end{align*}

\noindent
$\bullet$ {\bf{Qualitative PCR}}:\\

Applying the proportional conflict redistribution for the first partial conflict $qm_1(A)qm_2(B)qm_3(B\cup C)$, one gets:
%
\begin{align*}
\frac{x_{1,A}}{L_2}& =\frac{y_{1,B}}{L_3}=\frac{z_{1,B\cup C}}{L_5}=\frac{L_2L_3L_5}{L_2+L_3+L_5} =\frac{L_{\frac{5}{6}}}{L_{10}}=L_{(\frac{5}{6}\div 10)\cdot 6}=L_{\frac{3}{6}}
\end{align*}
\noindent
so,
$$x_{1,A}=L_2L_{\frac{3}{6}} = L_{\frac{2\cdot 3}{6\cdot 6}}= L_{\frac{1}{6}}$$
$$y_{1,B}=L_3L_{\frac{3}{6}} = L_{\frac{3\cdot 3}{6\cdot 6}}= L_{\frac{1.5}{6}}$$
$$z_{1,B\cup C}=L_5L_{\frac{3}{6}} = L_{\frac{5\cdot 3}{6\cdot 6}}= L_{\frac{2.5}{6}}$$

\noindent
Applying the proportional conflict redistribution for the second partial conflict $qm_1(A)qm_2(B)qm_3(A\cup B\cup C \cup D)$, one gets:
%
\begin{align*}
\frac{x_{2,A}}{L_2}& =\frac{y_{2,B}}{L_3}=\frac{w_{1,A\cup B\cup C \cup D}}{L_1}=\frac{L_2L_3L_1}{L_2+L_3+L_1} =\frac{L_{\frac{1}{6}}}{L_{6}}=L_{(\frac{1}{6}\div 6)\cdot 6}=L_{\frac{1}{6}}
\end{align*}
\noindent
so,
$$x_{2,A}=L_2L_{\frac{1}{6}} = L_{\frac{2\cdot 1}{6\cdot 6}}= L_{\frac{1/3}{6}}$$
$$y_{2,B}=L_3L_{\frac{1}{6}} = L_{\frac{3\cdot 1}{6\cdot 6}}= L_{\frac{1/2}{6}}=L_{\frac{0.5}{6}}$$
$$w_{1,A\cup B\cup C \cup D}=L_1L_{\frac{1}{6}} = L_{\frac{1\cdot 1}{6\cdot 6}}= L_{\frac{1/6}{6}}$$

\noindent
Applying the proportional conflict redistribution for the third partial conflict $qm_1(A)qm_2(A\cup B\cup C \cup D)qm_3(B\cup C)$, one gets:
\begin{align*}
\frac{x_{3,A}}{L_2}& =\frac{w_{2,A\cup B\cup C \cup D}}{L_3}=\frac{z_{2,B\cup C}}{L_5}=\frac{L_2L_3L_5}{L_2+L_3+L_5}=L_{\frac{3}{6}}
\end{align*}
\noindent
so,
$$x_{3,A}=L_2L_{\frac{3}{6}} = L_{\frac{1}{6}}$$
$$w_{2,A\cup B\cup C \cup D}=L_3L_{\frac{3}{6}} =L_{\frac{1.5}{6}}$$
$$z_{2,B\cup C }=L_5L_{\frac{3}{6}} =L_{\frac{2.5}{6}}$$

\noindent
Summing, we get:
%
\begin{align*}
qm_{\PCR}(A)&   =  L_{\frac{1}{6}} + L_{\frac{1}{6}} + L_{\frac{1/3}{6}} + L_{\frac{1}{6}}  = L_{\frac{10/3}{6}}\\
qm_{\PCR}(B) &  =  L_{\frac{12}{6}} + L_{\frac{1.5}{6}} + L_{\frac{0.5}{6}}  = L_{\frac{14}{6}}\\
qm_{\PCR}(B\cup C) &  =  L_{\frac{10}{6}} + L_{\frac{2.5}{6}} + L_{\frac{2.5}{6}} = L_{\frac{15}{6}}\\
qm_{\PCR}(A\cup B\cup C \cup D) &  =  L_{\frac{2}{6}} + L_{\frac{1/6}{6}} + L_{\frac{1.5}{6}}  = L_{\frac{22/6}{6}}
\end{align*}

\noindent
We can check that $qm_{\PCR}(.)$ is quasi-normalized without approximations (\emph{i.e.} when working within the refined set of linguistic labels by keeping fractional indexes), but it is not quasi-normalized when using approximations of fractional indexes if we want to work back within the original set of linguistic labels $L=\{L_{\min}=L_0,L_1,L_2,L_3,L_4,L_5,L_{\max}=L_6\}$.\\

\noindent
$\bullet$ {\bf{Discounted Qualitative PCR}} (formula \eqref{DPCR}):\\

\noindent
Let's consider the discounting factor $\alpha=0.6$. Consider the previous example and discount it according to \eqref{DPCR} applied in the qualitative domain. One obtains:
%
\begin{align*}
qm_{\DPCR}(A) & = L_{\frac{1}{6}} + 0.6 \cdot (L_{\frac{1}{6}} + L_{\frac{1/3}{6}} + L_{\frac{1}{6}})  = L_{\frac{1}{6}} + 0.6 \cdot L_{\frac{7/3}{6}} = L_{\frac{2.4}{6}}
\end{align*}
\begin{align*}
qm_{\DPCR}(B) & = L_{\frac{12}{6}} + 0.6 \cdot (L_{\frac{1.5}{6}} + L_{\frac{0.5}{6}})  = L_{\frac{1}{6}} + 0.6 \cdot L_{\frac{2}{6}} = L_{\frac{13.2}{6}}
\end{align*}
\begin{align*}
qm_{\DPCR}(B\cup C) & = L_{\frac{10}{6}} + 0.6 \cdot (L_{\frac{2.5}{6}} + L_{\frac{2.5}{6}})  = L_{\frac{10}{6}} + 0.6 \cdot L_{\frac{5}{6}} = L_{\frac{13}{6}}
\end{align*}
\begin{align*}
qm_{\DPCR}(A\cup B\cup C) & = (1-\alpha)qm_1(A)qm_2(B) qm_2(B\cup C) = 0.4 \cdot L_2L_3L_5= 0.4 \cdot L_{\frac{5}{6}} = L_{\frac{2}{6}}
\end{align*}
\begin{align*}
qm_{\DPCR}(A\cup B\cup C\cup D) & = L_{\frac{2}{6}}  + 0.6\cdot(L_{\frac{1/6}{6}} + L_{\frac{1.5}{6}}) + 0.4\cdot(L_2L_3L_1 + L_2L_3L_5)\\
& = L_{\frac{2}{6}} + 0.6 \cdot L_{\frac{5/3}{6}} + 0.4\cdot (L_{\frac{1}{6}} + L_{\frac{5}{6}})\\
& = L_{\frac{2}{6}} + L_{\frac{1}{6}} +  0.4\cdot L_{\frac{6}{6}} = L_{\frac{3}{6}} + L_{\frac{2.4}{6}}= L_{\frac{5.4}{6}}
\end{align*}

\noindent
$qm_{\DPCR}(.)$ is quasi-normalized without approximations, but it is not with approximations.

\section{Conclusions}
\label{conclusion}
\label{Sec7}

With the recent development of qualitative methods for reasoning under uncertainty developed in Artificial Intelligence, more and more experts and scholars have expressed great interest on qualitative information fusion, especially those working in the development of modern multi-source systems for defense, robot navigation, mapping, localization and path planning and so on. In this paper, we propose some solutions to handle the conflict and to weight the imprecision of the responses of the experts, from the classical combination rules for qualitative and quantitative beliefs. Hence, we have presented a mixed rule given by a weighted sum of the conjunctive and disjunctive rules. The weights are defined from a measure of non-specifity calculated by the cardinality of the responses of the experts. This rule transfers the partial conflict on partial ignorance. Again, the proportional conflict distribution rule redistributes the partial conflict on the element implied in this conflict. We propose an extension of this rule by a discounting procedure, thereby, a part of the partial conflict is also redistributed on the partial ignorance. We have introduced a measure of conflict between pair of experts and another measure of non-conflict between pair of experts, as to quantify this part. In order to take heed of the non-specifity and to redistributed the partial conflict, we propose a fused rule of these two new rules. This rule is created in such way that we retain a kind of continuity of the mass on the partial ignorance, between both cases with and without partial conflict. Illustrating examples have been presented in detail to explain how the new rules work for quantitative and qualitative beliefs. The study of these new rules shows that the classical combination rules in the belief functions theory cannot take precisely into account the non-specifity of the experts and the partial conflict of the experts. This is specially important for qualitative belief.

\bibliographystyle{IEEEtran}

\end{document}